\documentclass[nohyperref]{article}

\usepackage{amsopn,amsmath,amsthm,amssymb}
\usepackage{wrapfig}
\usepackage{multirow}
 \usepackage{mathtools}

\usepackage{anyfontsize}

\usepackage{color}
\usepackage{tikz}
\usetikzlibrary{arrows,shapes,snakes,automata,backgrounds,fit,petri}
\usepackage{adjustbox}

\newcommand{\RN}[1]{%
	\textup{\lowercase\expandafter{\it \romannumeral#1}}%
}

\makeatletter
\newcommand{\distas}[1]{\mathbin{\overset{#1}{\kern\z@\sim}}}%

\usepackage{enumitem}

\usepackage{algorithm}
\usepackage{algorithmic}

\newcommand{\beq}{\vspace{0mm}\begin{equation}}
\newcommand{\eeq}{\vspace{0mm}\end{equation}}
\newcommand{\beqs}{\vspace{0mm}\begin{eqnarray}}
\newcommand{\eeqs}{\vspace{0mm}\end{eqnarray}}
\newcommand{\barr}{\begin{array}}
\newcommand{\earr}{\end{array}}

\newcommand{\R}{\mathbb{R}}

\newcommand{\Lcal}{\mathcal{L}}

\newcommand{\Ocal}{\mathcal{O}}

\newtheorem{theorem}{Theorem} 

\DeclareMathOperator{\RR}{\mathbb{R}} 

\ifx\assumption\undefined
\newtheorem{assumption}{Assumption}
\fi

\ifx\definition\undefined
\newtheorem{definition}{Definition}
\fi

\ifx\remark\undefined
\newtheorem{remark}{Remark}
\fi
\newcommand{\norm}[1]{\left\| #1 \right\|}

\newcommand{\Abs}[1]{\left| #1 \right| }

\newcommand{\PVar}[1]{\operatorname{Var}\left[#1\right]}
\newcommand{\PExv}[1]{\mathbf{E}\left[#1\right]}
\usepackage{arydshln}
\usepackage{microtype}
\usepackage{graphicx}
\usepackage{subfigure}
\usepackage{booktabs} 

\usepackage{hyperref}

\usepackage[accepted]{icml2022}

\usepackage{amsmath}
\usepackage{amssymb}
\usepackage{mathtools}
\usepackage{amsthm}

\usepackage[capitalize,noabbrev]{cleveref}

\theoremstyle{plain}
\usepackage[textsize=tiny]{todonotes}

\icmltitlerunning{Low-Precision Stochastic Gradient Langevin Dynamics}

\begin{document}

\twocolumn[
\icmltitle{Low-Precision Stochastic Gradient Langevin Dynamics}




\begin{icmlauthorlist}
\icmlauthor{Ruqi Zhang}{yyy}
\icmlauthor{Andrew Gordon Wilson}{comp}
\icmlauthor{Christopher De Sa}{sch}
\end{icmlauthorlist}

\icmlaffiliation{yyy}{The University of Texas at Austin}
\icmlaffiliation{comp}{New York University}
\icmlaffiliation{sch}{Cornell University}

\icmlcorrespondingauthor{Ruqi Zhang}{ruqiz@utexas.edu}

\icmlkeywords{Machine Learning, ICML}

\vskip 0.3in
]



\printAffiliationsAndNotice{}  

\begin{abstract}
While low-precision optimization has been widely used to accelerate deep learning, low-precision sampling remains largely unexplored. As a consequence, sampling is simply infeasible in many large-scale scenarios, despite providing remarkable benefits to generalization and uncertainty estimation for neural networks. In this paper, we provide the first study of low-precision Stochastic Gradient Langevin Dynamics (SGLD), showing that its costs can be significantly reduced without sacrificing performance, due to its intrinsic ability to handle system noise. We prove that the convergence of low-precision SGLD with full-precision gradient accumulators is less affected by the quantization error than 
its SGD counterpart in the strongly convex setting. To further enable low-precision gradient accumulators, we develop a new quantization function for SGLD that preserves the variance in each update step. We demonstrate that low-precision SGLD achieves comparable performance to full-precision SGLD with only 8 bits on a variety of deep learning tasks.
\end{abstract}

\section{Introduction}

Low-precision optimization has become increasingly popular in reducing computation and memory costs of training deep neural networks (DNNs). It uses fewer bits to represent numbers in model parameters, activations, and gradients, and thus can drastically lower resource demands~\citep{gupta2015deep,zhou2016dorefa,de2017understanding,li2017training}. Prior work has shown that using 8-bit numbers in training DNNs achieves about 4$\times$ latency speed ups and memory reduction compared to 32-bit numbers on a wide variety of deep learning tasks~\citep{sun2019hybrid,yang2019swalp,wang2018training,banner2018scalable}.
As datasets and architectures grow rapidly, performing low-precision optimization enables training large-scale DNNs efficiently and enables many applications on different hardware and platforms.

Despite the impressive progress in low-precision optimization, low-precision sampling remains largely unexplored. However, we believe stochastic gradient Markov chain Monte Carlo (SGMCMC) methods~\citep{welling2011bayesian,chen2014stochastic,ma2015complete} are particularly suited for low-precision arithmetic because of their intrinsic robustness to system noise. In particular: (1) SGMCMC explores weight space instead of converging to a single point, thus it should not require precise  weights or gradients; (2) SGMCMC even adds noise to the system to encourage exploration and so is naturally more tolerant to quantization noise; (3) SGMCMC performs Bayesian model averaging during testing using an ensemble of models, which 
allows coarse representations of individual models to be compensated by the overall model average~\citep{zhu2019binary}. 

SGMCMC is particularly compelling in Bayesian deep learning due to its ability to characterize complex and multimodal DNN posteriors, providing state-of-the-art generalization accuracy and calibration~\citep{zhang2019cyclical,li2016learning,gan2016scalable,heek2019bayesian}. Moreover, low-precision approaches are especially appealing in this setting, where at test time we must store samples from a posterior over millions of parameters, and perform multiple forward passes through the corresponding models, which incurs significant memory and computational expenses.

In this paper, we give the first comprehensive study of low-precision Stochastic Gradient Langevin Dynamics (SGLD)~\citep{welling2011bayesian}, providing both theoretical convergence bounds and promising empirical results in deep learning. On strongly log-concave distributions (i.e. strongly convex functions for SGD), we prove that the convergence of SGLD with full-precision gradient accumulators is more robust to the quantization error than its counterpart in SGD. Surprisingly, we find that SGLD with low-precision gradient accumulators can diverge arbitrarily far away from the target distribution with small stepsizes. We identify the source of the issue and develop a new quantization function to correct the bias with minimal overhead. Empirically, we demonstrate low-precision SGLD across different tasks, showing that it is able to provide superior generalization and uncertainty estimation using just 8 bits. We summarize our contributions as follows:

\begin{itemize}
    \item We provide a methodology for SGLD to leverage low-precision computation, including a new quantization function, while still guaranteeing its convergence to the target distribution. 
    \item We offer theoretical results which explicitly show how quantization error affects the convergence of the sampler in the strongly convex setting, proving its robustness to quantization noise over SGD. 
    \item We show SGLD is particularly suitable for low-precision deep learning over a range of experiments, including logistic regression, Bayesian neural networks on image and text classification.
\end{itemize}
In short, low-precision SGLD is often a compelling alternative to standard SGLD, improving speed and memory efficiency, while retaining accuracy. Moreover, SGLD is arguably more amenable to low-precision computations than SGD. Our code is \href{https://github.com/ruqizhang/low-precision-sgld}{\underline{available here}}.

\section{Related Work}

To speed up SGLD training, most existing work is on distributed learning with synchronous or asynchronous communication~\citep{ahn2014distributed,chen2016stochastic,li2019communication}. Another direction is to shorten training time by accelerating the convergence using variance reduction techniques~\citep{dubey2016variance,baker2019control}, importance sampling~\citep{deng2020contour} or a cyclical learning rate schedule~\citep{zhang2019cyclical}. To speed up SGLD during testing, distillation techniques are often used to save both compute and memory which transfer the knowledge of an ensemble of models to a single model~\citep{korattikara2015bayesian,wang2018adversarial}. 

Low-precision computation has become one of the most common approaches to reduce latency and memory consumption in deep learning and is widely supported on new emerging chips including CPUs, GPUs and TPUs~\citep{micikevicius2017mixed,krishnamoorthi2018quantizing,esser2019learned}. Two main directions to improve low-precision training include developing new number formats~\citep{sun2019hybrid,sun2020ultra} or studying mixed-precision schemes~\citep{courbariaux2015binaryconnect,zhou2016dorefa,banner2018scalable}. Recently, one line of work applies the Bayesian framework to learn a deterministic quantized neural network~\citep{soudry2014expectation,cheng2015training,achterhold2018variational,van2020bayesian,meng2020training}.

Low-precision computation is largely unexplored for Bayesian neural networks, despite their specific promise in this domain. \citet{su2019sampling} proposes a method to train binarized variational BNNs, and \citet{cai2018vibnn} develops efficient hardware for training low-precision variational BNNs. The only work on low-precision MCMC known to us is \citet{ferianc2021effects}, which directly applies \emph{post-training} quantization techniques from optimization~\citep{jacob2018quantization} to convert BNNs trained by Stochastic Gradient Hamiltonian Monte Carlo~\citep{chen2014stochastic} into low-precision models. We instead study training low-precision models by SGLD from scratch, to accelerate 
both training and testing.

\section{Preliminaries}
\subsection{Stochastic Gradient Langevin Dynamics}
In the Bayesian setting, given some dataset $\mathcal{D}$, a model with parameters $\theta\in\RR^d$, and a prior $p(\theta)$, we are interested in sampling from the posterior $p(\theta|\mathcal{D})\propto\exp(-U(\theta))$, where the energy function is
\[
U(\theta) = -\sum_{x\in\mathcal{D}}\log p(x|\theta)-\log p(\theta).
\]
When the dataset is large, the cost of computing a sum over the entire dataset is expensive. Stochastic Gradient Langevin Dynamics (SGLD)~\citep{welling2011bayesian} reduces the cost by using a stochastic gradient estimation $\nabla\tilde{U}$, which is an unbiased estimator of $\nabla U$ based on a subset of the dataset $\mathcal{D}$. Specifically, SGLD updates the parameter $\theta$ in the $(k+1)$-th step following the rule
\begin{align}
    \theta_{k+1} = \theta_{k} -\alpha\nabla\tilde{U}(\theta_{k}) + \sqrt{2\alpha}\xi_{k+1},\label{eq:sgld-update}
\end{align}
where $\alpha$ is the stepsize and $\xi$ is a standard Gaussian noise. Compared to the SGD update, the only difference is that SGLD adds an additional Gaussian noise in each step, which essentially enables SGLD to characterize the full distribution instead of converging to a single point. The close connection between SGLD and SGD makes it convenient to implement and run on existing deep learning tasks for which SGD is the typical learning algorithm.

\subsection{Low-Precision Training}\label{sec:lp-training}
We study training a low-precision model by SGLD from scratch, to reduce both training and testing costs. Specifically, we follow the framework in prior work to quantize the weights, activations, backpropagation errors, and gradients~\citep{wu2018training,wang2018training,yang2019swalp}. We mainly consider the effect of weight and gradient quantization following previous work~\citep{li2017training,yang2019swalp}. Please refer to Appendix~\ref{sec:lp-formulation} for more details.

\subsubsection{Number Representations}
To represent numbers in low-precision, one simple way is to use \emph{fixed point}, which has been utilized in both theory and practice ~\citep{gupta2015deep,lin2016fixed,li2017training,yang2019swalp}. Specifically, suppose that we use $W$ bits to represent numbers with $F$ of those $W$ bits to represent the fractional part. Then there is a distance between consecutive representable numbers, $\Delta=2^{-F}$, which is called \emph{quantization gap}. The representable numbers also have a lower bound $l$ and an upper bound $u$, where
$$l = -2^{W-F-1}, ~~~u = 2^{W-F-1}-2^{-F}.$$ 
As shown above, when the number of bits decreases, the accuracy of number representation decreases.  
We use this type of number representation in our theoretical analysis and empirical demonstration following previous work~\citep{li2017training,yang2019swalp}.

Another common type of number representation is \emph{floating point} where each number has its own exponent part. Between fixed point and floating point, there is \emph{block floating point} which allows all numbers within a block to share the same exponent~\citep{song2018computation}. We use block floating point for deep learning experiments since it has been shown more favorable for deep models~\citep{yang2019swalp}.

\subsubsection{Quantization}
Having low-precision number representation in hand, we also need a quantization function $Q$ to convert a real-valued number into a low-precision number. Such functions include \emph{deterministic rounding} and \emph{stochastic rounding}. Particularly, the deterministic rounding function $Q^d$ quantizes a number to its nearest representable neighbor as follows:
\[
Q^d(\theta) = \text{sign}(\theta)\cdot \text{clip}\left(\Delta\left\lfloor\frac{\Abs{\theta}}{\Delta}+\frac{1}{2}\right\rfloor, l, u\right),
\]
where $\text{clip}(x,l,u) = \max(\min(x,u),l)$.
Instead, stochastic rounding $Q^s$ quantizes a number with a probability based on the distance to its representable neighbor:
\begin{align*}
Q^s(\theta) =
    \begin{cases}
      \text{clip}\left(\Delta\left\lfloor\frac{\theta}{\Delta}\right\rfloor, l, u\right), &\text{w.p. } \left\lceil\frac{\theta}{\Delta}\right\rceil -  \frac{\theta}{\Delta} \\
      \text{clip}\left(\Delta\left\lceil\frac{\theta}{\Delta}\right\rceil, l, u\right), &\text{w.p. } 1-\left(\left\lceil\frac{\theta}{\Delta}\right\rceil -  \frac{\theta}{\Delta}\right).
    \end{cases}
\end{align*}
    
An important property of $Q^s$ is that $\mathbf{E}[Q^s(\theta)]=\theta$, which means the quantized number is unbiased. $Q^s$ is generally preferred over $Q^d$ in practice since it can preserve gradient information especially when the gradient update is smaller than the quantization gap~\citep{gupta2015deep,wang2018training}. In what follows, we use $Q_W$ and $\Delta_W$ to denote the weights' quantizer and quantization gap, $Q_G$ and $\Delta_G$ to denote gradients' quantizer and quantization gap.

To do the gradient update in low-precision training, there are two common choices depending on whether we store an additional copy of full-precision weights.
\emph{Full-precision gradient accumulators} use a full-precision weight buffer to accumulate gradient updates and only quantize weights before computing gradients. SGD with full-precision gradient accumulators (SGDLP-F) updates the weights as the following,
\[
\theta_{k+1} = \theta_{k} - \alpha Q_G\left(\nabla\tilde{U}(Q_W\left(\theta_{k})\right)\right),
\]
where we use full-precision $\theta_{k+1}$ and $\theta_k$ in the update, and only quantize the weight for forward and backward propagation~\citep{courbariaux2015binaryconnect,li2017training}.

However, gradient accumulators have to be frequently updated during training, therefore it will be ideal to also represent it in low-precision to further reduce the costs. To achieve it, we could instead do the update as follows,
\begin{align}
    \theta_{k+1} = Q_W\left(\theta_{k} - \alpha Q_G\left(\nabla\tilde{U}(\theta_{k})\right)\right),\label{eq:sgdlp-l}
\end{align}
where $\theta$ is always represented in low-precision.
This update of SGD is called using \emph{low-precision gradient accumulators} (SGDLP-L). 
Both full- and low-precision gradient accumulators have been widely used: low-precision gradient accumulators are cheaper and faster because of having all numbers in low-precision, whereas full-precision gradient accumulators generally have better performance because of more precisely reflecting small gradient updates~\citep{courbariaux2015binaryconnect,li2017training}.

\section{Low-Precision SGLD }\label{sec:sgldlp}
In this section, we first study the convergence of low-precision SGLD with full-precision gradient accumulators (SGLDLP-F) on strongly log-concave distributions (i.e. the energy function is strongly convex) and show that SGLDLP-F is less affected by the quantization error than its SGD counterpart. Next we analyze low-precision SGLD with low-precision gradient accumulators (SGLDLP-L) under the same setup and prove that SGLDLP-L can diverge arbitrarily far away from the target distribution with a small stepsize, which however is typically required by SGLD to reduce asymptotic bias. Finally, we solve this problem by developing a variance-corrected quantization function and further prove that with this quantization function, SGLDLP-L converges with small stepsizes.

\subsection{Full-Precision Gradient Accumulators}
As shown in Equation~\eqref{eq:sgld-update}, the update of SGLD is simply a SGD update plus a Gaussian noise. Therefore the low-precision formulation for SGD in Section~\ref{sec:lp-training} can be naturally extended to SGLD training. Similar to SGDLP-F, we can do low-precision SGLD with full-precision gradient accumulators~(SGLDLP-F) as the following:
\begin{align}
    \theta_{k+1} = \theta_{k} - \alpha Q_G\left(\nabla\tilde{U}(Q_W\left(\theta_{k})\right)\right) + \sqrt{2\alpha}\xi_{k+1},\label{eq:highacc}
\end{align}
which can also be viewed as SGDLP-F plus a Gaussian noise in each step. However, the Gaussian noise turns out to help counter-effect the noise introduced by quantization, making SGLDLP-F more robust to inaccurate number representation and converging better than SGDLP-F as we show later in this section.  

We now prove the convergence of SGLDLP-F. Our analysis is built upon the 2-Wasserstein distance bounds of SGLD in~\citet{dalalyan2019user}, where the target distribution is assumed to be smooth and strongly log-concave. We additionally assume the energy function has Lipschitz Hessian following recent work in low-precision optimization~\citep{yang2019swalp}. In summary, the energy function $U$ has the following assumptions, $\forall~ \theta,\theta'\in \mathbb{R}^d$, it satisfies
\begin{align*}
\begin{cases}
      U(\theta) - U(\theta') - \nabla U(\theta')^\intercal (\theta - \theta') \ge (m/2) \norm{\theta - \theta'}_2^2, &  \\
      \norm{\nabla U(\theta) - \nabla U(\theta')}_2 \le M \norm{\theta - \theta'}_2, & \\
      \| \nabla^2 U(\theta) - \nabla^2 U(\theta') \|_2 \le \Psi \| \theta - \theta' \|_2, &
    \end{cases}
\end{align*}
for some positive constants $m$, $M$ and $\Psi$. We further assume that the variance of the stochastic gradient is bounded $\mathbf{E}[\| \nabla\tilde{U}(\theta) - \nabla U(\theta) \|_2^2 ]\le \kappa^2$ for some constant $\kappa$. For simplicity, we consider SGLD with a constant stepsize $\alpha$. We use stochastic rounding for quantizing both weights and gradients as it is generally better than deterministic rounding and has also been used in previous low-precision theoretical analysis~\citep{li2017training,yang2019swalp}.

\begin{theorem}\label{thm:highacc}
We run SGLDLP-F under the above assumptions and with a constant stepsize $\alpha \le 2/(m+M)$. Let $\pi$ be the target distribution, $\mu_0$ be the initial distribution and $\mu_K$ be the distribution obtained by SGLDLP-F after $K$ iterations, then the 2-Wasserstein distance is
  \begingroup\makeatletter\def\f@size{8}\check@mathfonts
\def\maketag@@@#1{\hbox{\m@th\large\normalfont#1}}
\begin{align*}
    W_2(\mu_K, \pi)&\le (1-\alpha m)^KW_2(\mu_0, \pi) + 1.65 (M/m)(\alpha d)^{1/2}\\
    &\hspace{-4em} + \min\left( \frac{ \Psi \Delta_W^2 d }{4m}, \frac{M \Delta_W \sqrt{d}}{2 m} \right) + \sqrt{\frac{(\Delta_G^2 + M^2 \Delta_W^2)\alpha d + 4\alpha \kappa^2}{4m}}.
\end{align*}\endgroup
\end{theorem}
This theorem shows that SGLDLP-F converges to the accuracy floor $\min\left( \frac{ \Psi \Delta_W^2 d }{4m}, \frac{M \Delta_W \sqrt{d}}{2 m} \right)$ given large enough number of iterations $K$ and small enough stepsize $\alpha$. Besides, if we further assume that the energy function is quadratic, that is $\Psi=0$, then SGLDLP-F converges to the target distribution asymptotically. This matches the result of SGDLP-F on a quadratic function which converges to the optimum asymptotically \citep{li2017training}. Our theorem also recovers the bound in \citet{dalalyan2019user} when the quantization gap is zero (we ignore $1.65M$ in the denominator in their bound for simplicity).

However, when the energy function is not quadratic, the convergence of SGLDLP-F to the target distribution has a $\Ocal(\Delta_W^2)$ rate whereas SGDLP-F to the optimum has a $\Ocal(\Delta_W)$ rate~\citep{yang2019swalp}\footnote{Their bound $\mathcal{O}(\Delta_W^2)$ is stated for the squared norm therefore we take its square root to compare with our $W_2$ distance bound which is stated for the norm.}. Recall that $\Delta_W = 2^{-F}$ where $F$ is the number of fractional bits, our result suggests that asymptotically, SGLD only needs half the number of bits as SGD needs to achieve the same convergence accuracy! Our comparison between SGLD and SGD also fits into the literature in comparing sampling and optimization convergence bounds~\citep{ma2019sampling,talwar2019computational} (see Appendix~\ref{sec:bound_compare} for more details). In summary, our theorem implies how sensitive SGLD is to the quantization error, and actually suggests that sampling methods are more suitable with low-precision computation than optimization methods.

\begin{figure*}
	\vspace{-0mm}\centering
	\begin{tabular}{ccccc}		
		\hspace{-4mm}
		\includegraphics[width=3.5cm]{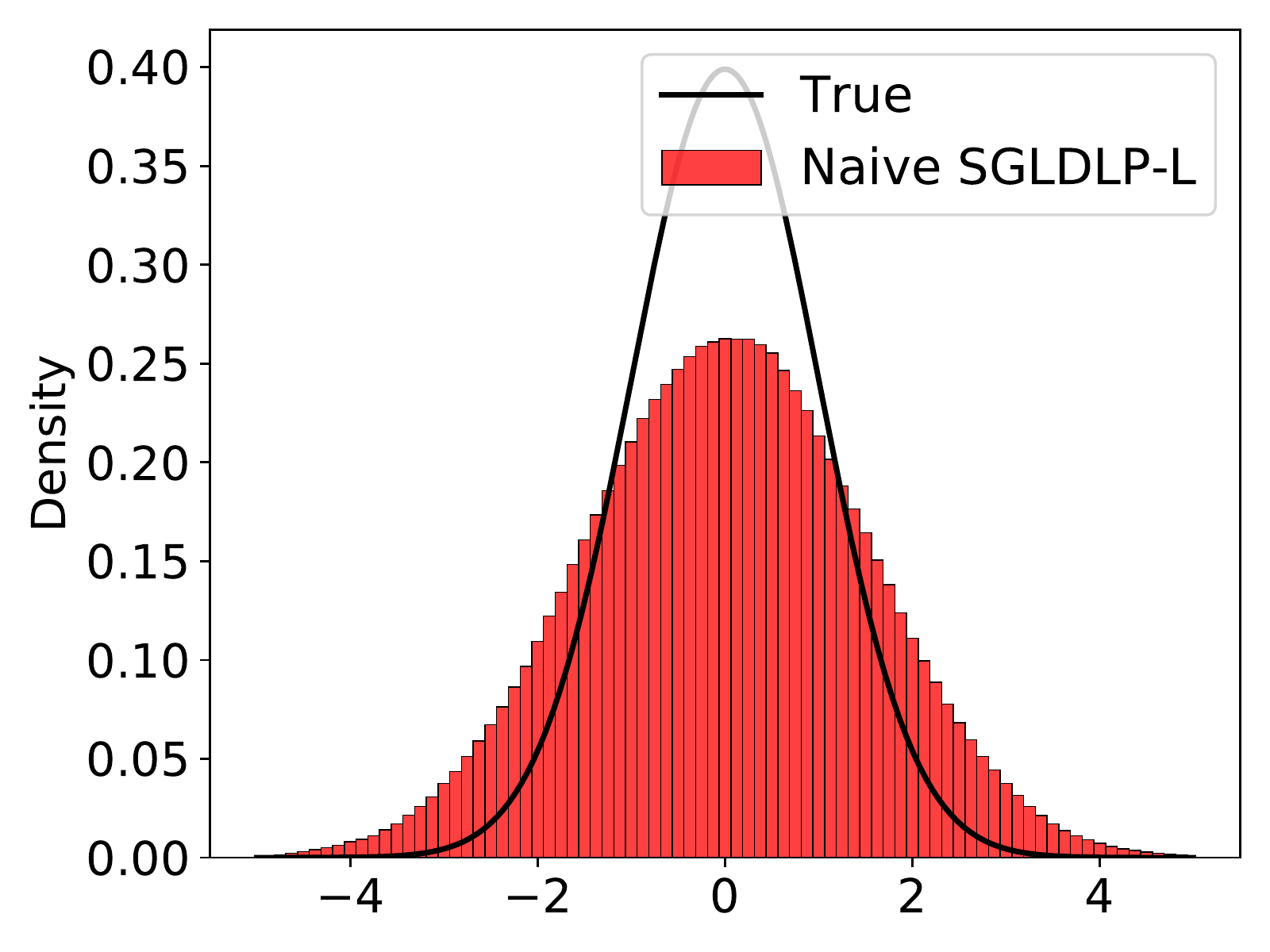}  &
        \hspace{-4mm}
        \includegraphics[width=3.5cm]{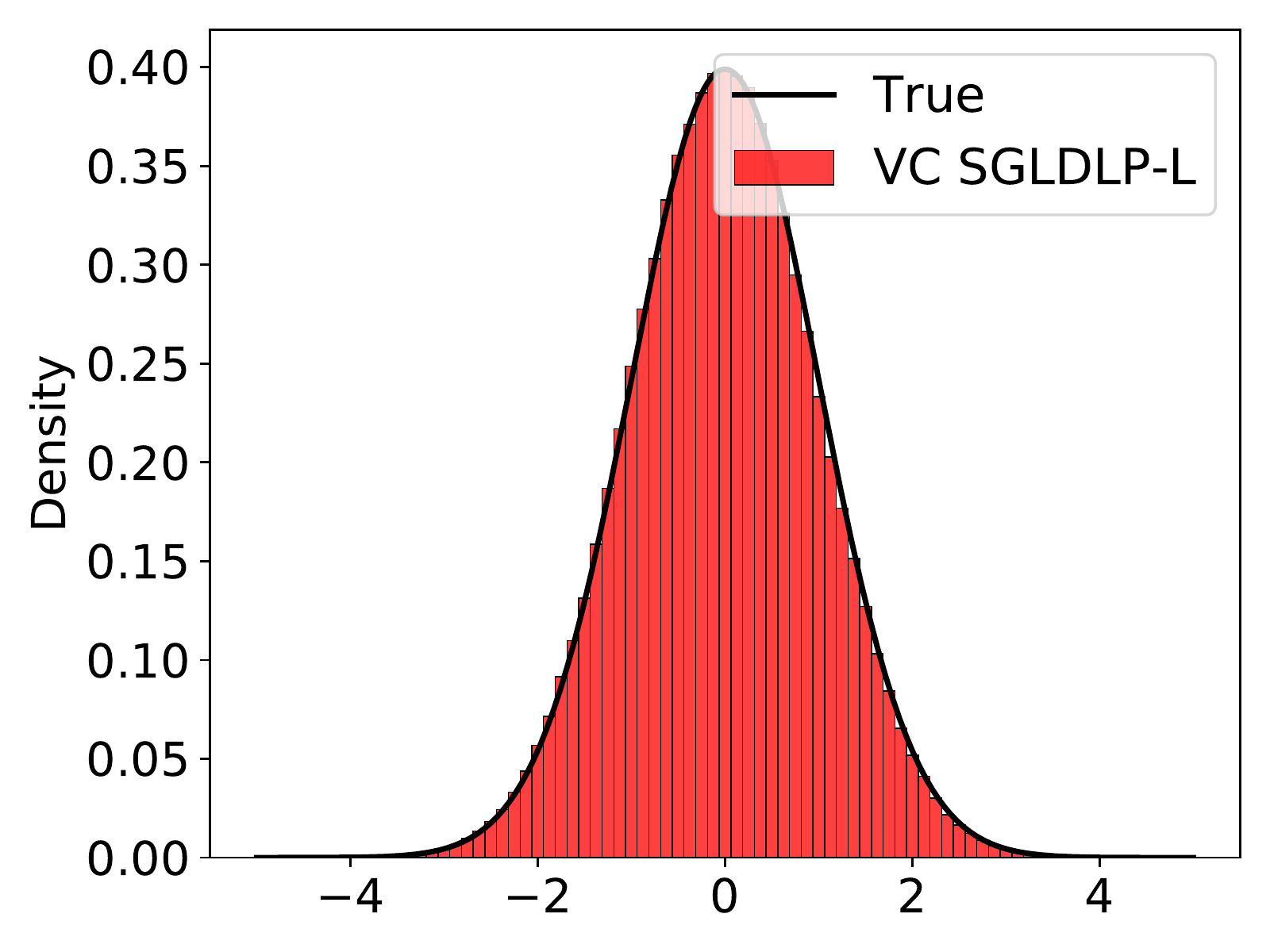}  &
        \hspace{-4mm}
        \includegraphics[width=3.5cm]{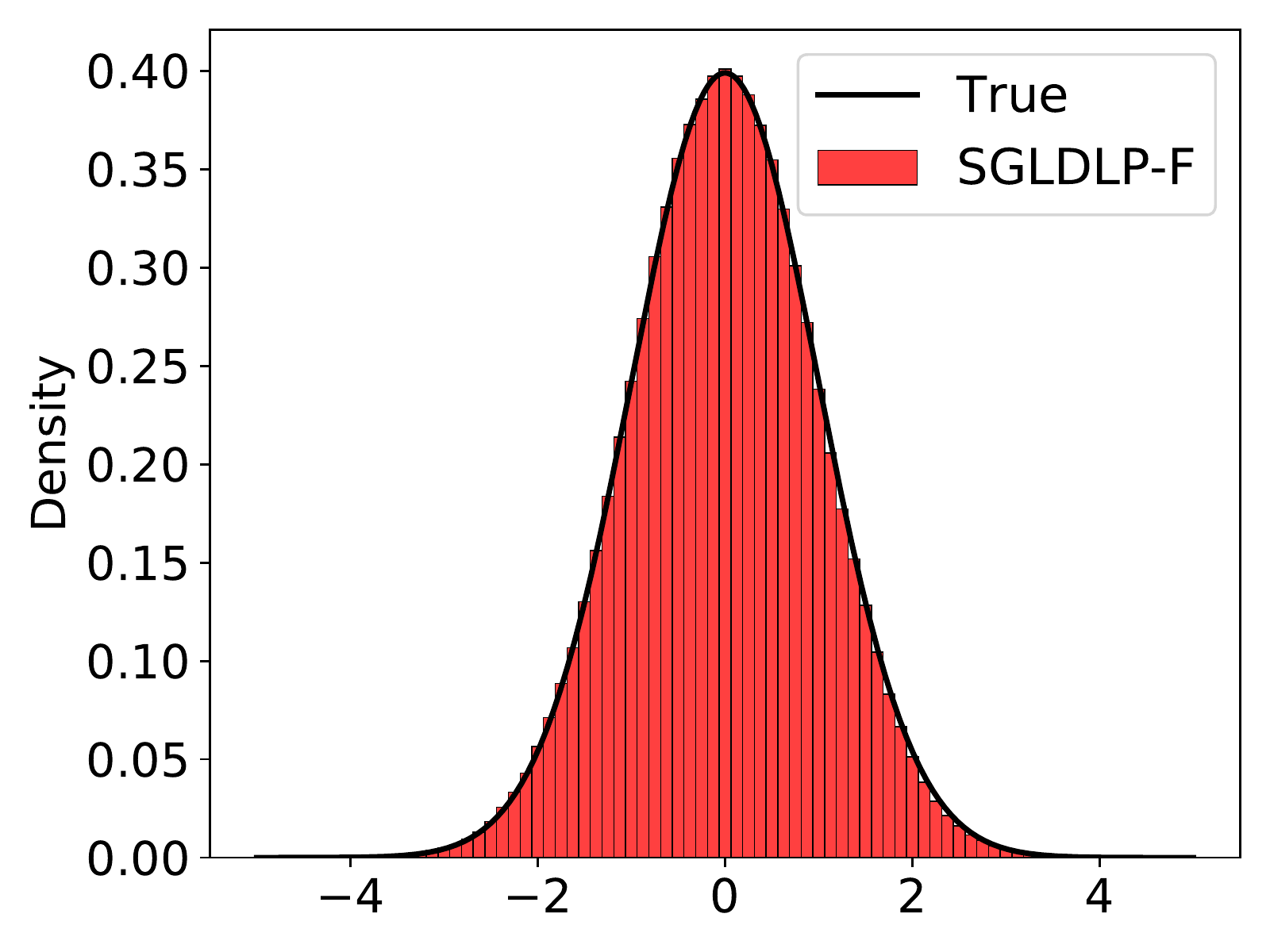}
        \\
		&(a) SGLD with $\alpha=0.001$	\\
		\hspace{-4mm}
		\includegraphics[width=3.5cm]{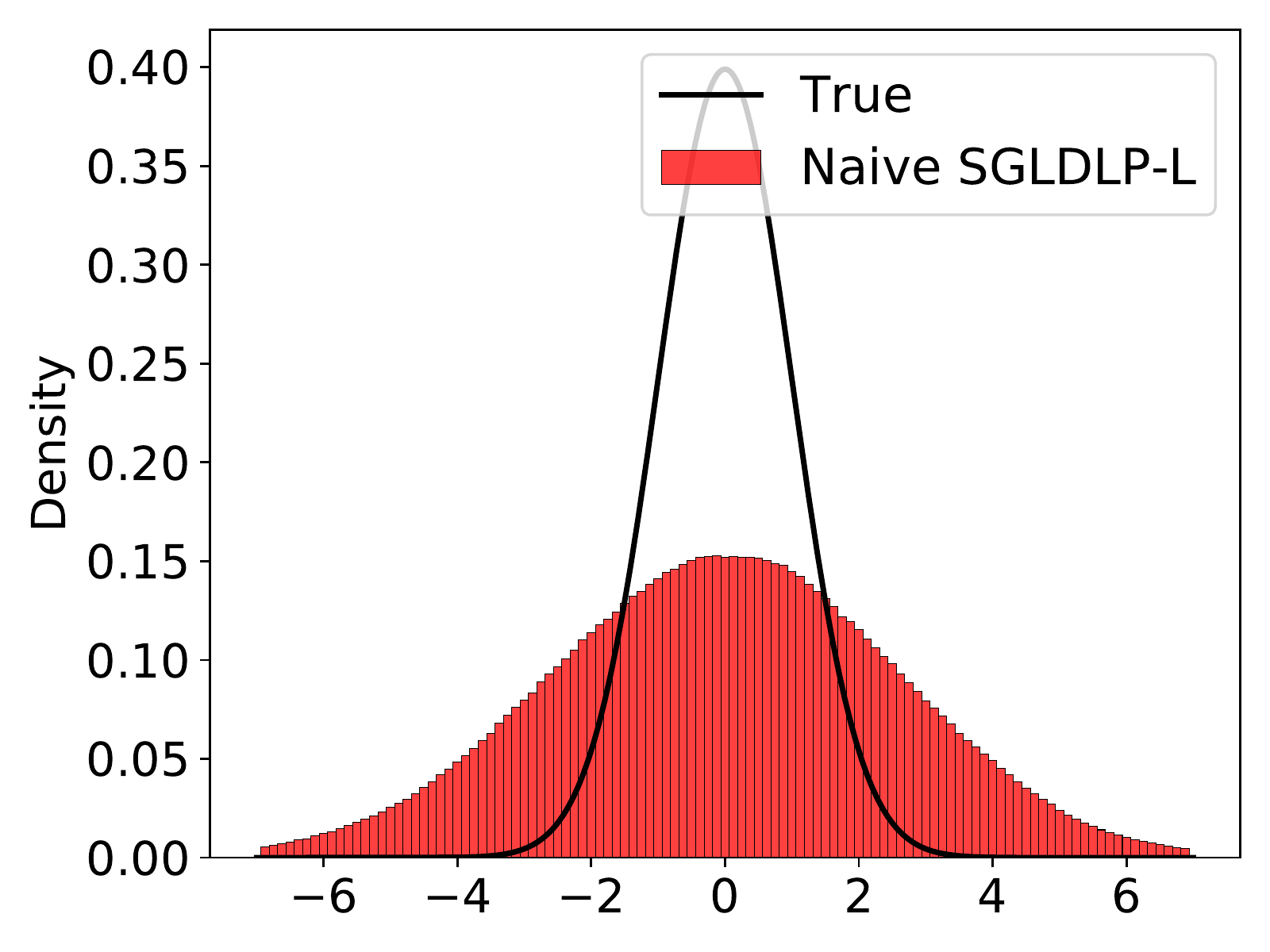}  &
        \hspace{-4mm}
        \includegraphics[width=3.5cm]{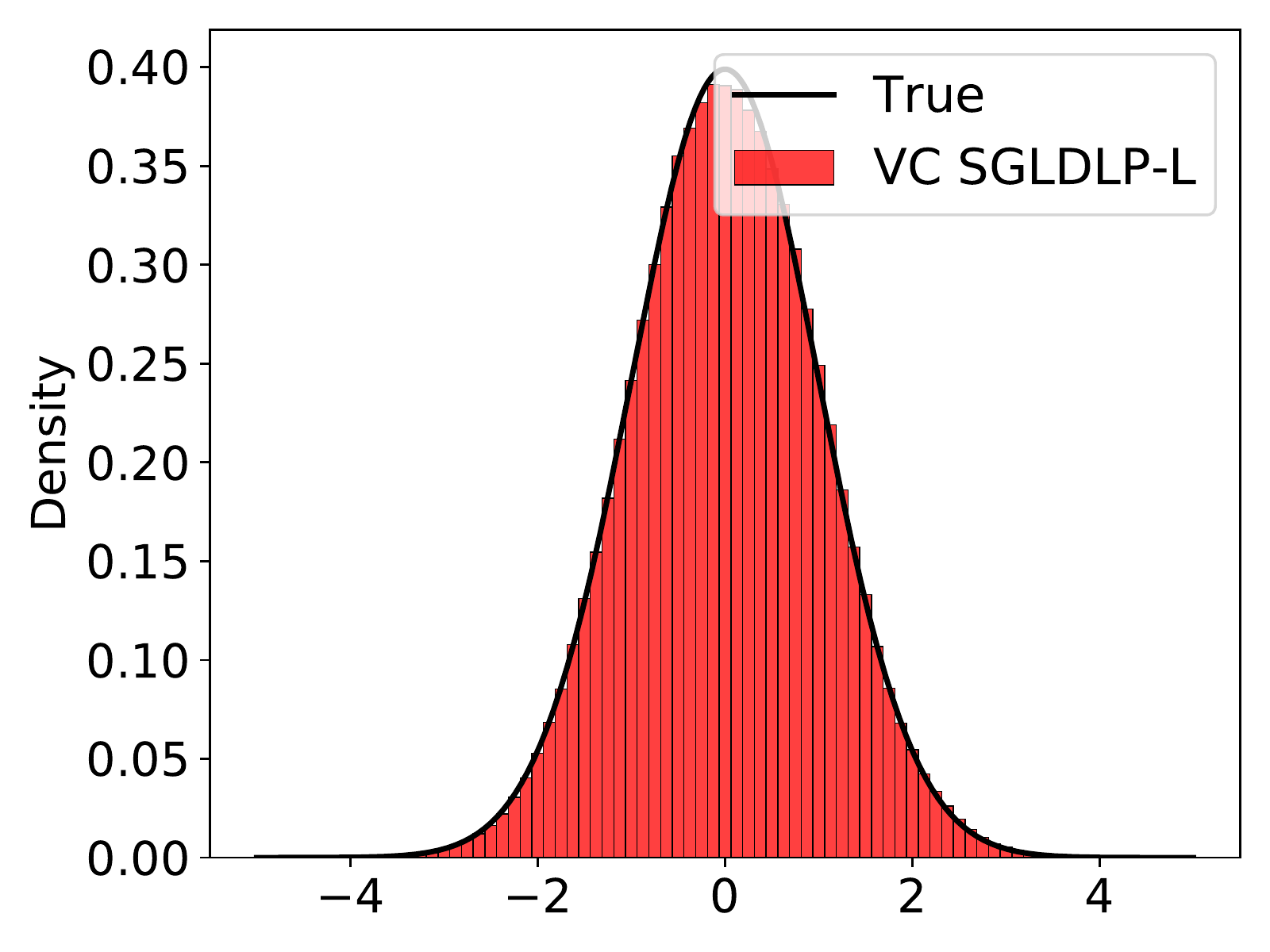}  &
        \hspace{-4mm}
        \includegraphics[width=3.5cm]{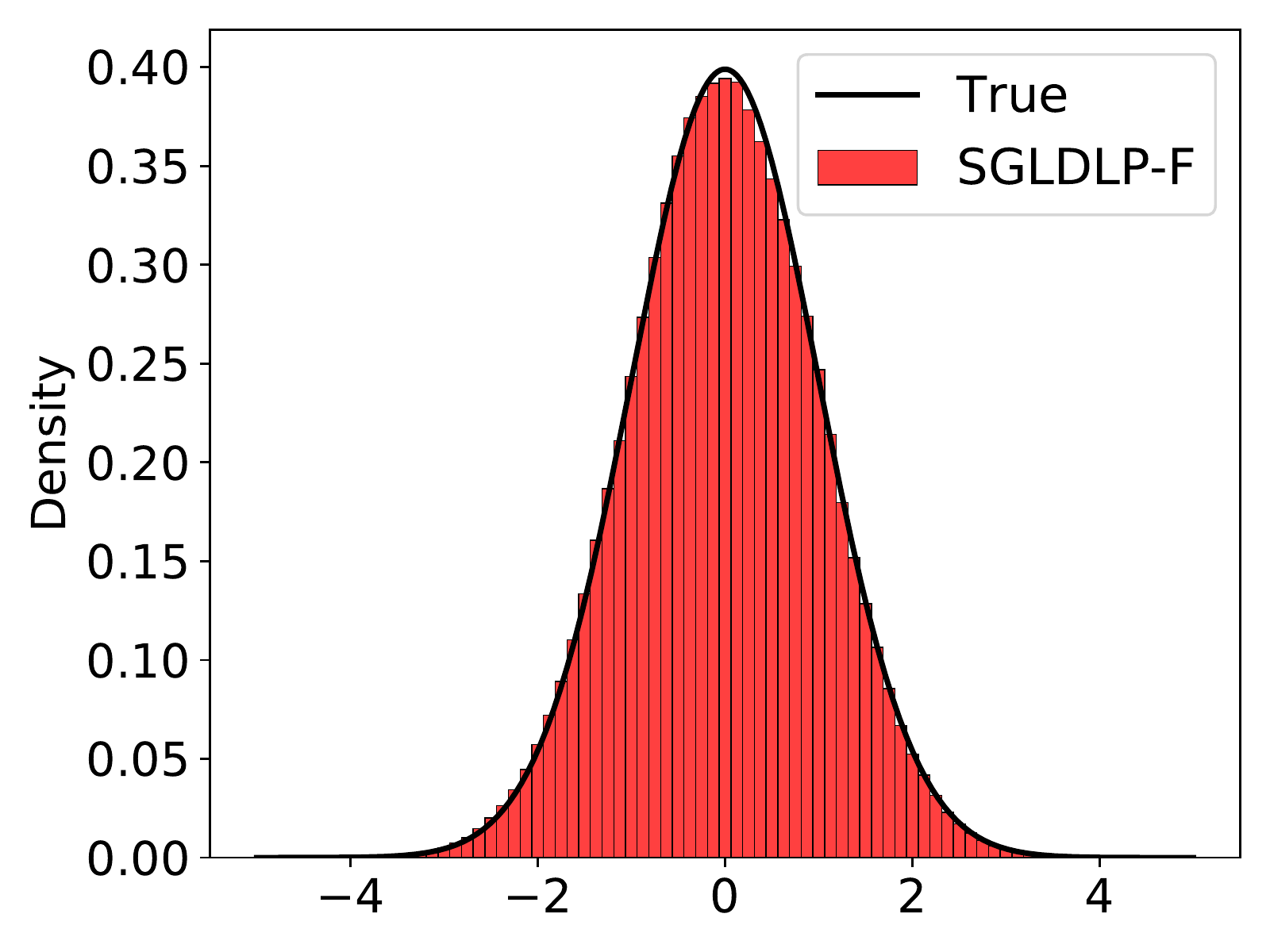}
        \\
		&(b) SGLD with $\alpha=0.0001$	\\

	\end{tabular}
	\caption{Low-precision SGLD with varying stepsizes on a Gaussian distribution. Variance-corrected SGLD with low-precision gradient accumulators (VC SGLDLP-L) and SGLD with full-precision gradient accumulators (SGLDLP-F) converge to the true distribution, whereas na\"ive SGLDLP-L diverges and the divergence increases as the stepsize decreases. }
	\label{fig:gaussian}
\end{figure*}

\subsection{Low-Precision Gradient Accumulators}
As mentioned before, it will be ideal to further reduce the costs using low-precision gradient accumulators. Mimicking the update of SGDLP-L in Equation~\eqref{eq:sgdlp-l}, it is natural to get the following update rule for SGLD with low-precision gradient accumulators (SGLDLP-L),
\begin{align}
    \theta_{k+1} = Q_W\left(\theta_{k} - \alpha Q_G\left(\nabla\tilde{U}(\theta_{k})\right) + \sqrt{2\alpha }\xi_{k+1}\right).\label{eq:lowacc}
\end{align}
Surprisingly, while we can prove a convergence result for SGLDLP-L, our theory and empirical results suggest that it can diverge arbitrarily far away from the target distribution with small stepsizes.
\begin{theorem}\label{thm:lowacc}
We run SGLDLP-L under the same assumptions as in Theorem~\ref{thm:highacc}. Let $\mu_0$ be the initial distribution and $\mu_K$ be the distribution obtained by SGLDLP-L after $K$ iterations, then
  \begingroup\makeatletter\def\f@size{8}\check@mathfonts
\def\maketag@@@#1{\hbox{\m@th\large\normalfont#1}}
\begin{align*}
    W_2(\mu_K, \pi)&\le (1-\alpha m)^KW_2(\mu_0, \pi) + 1.65 (M/m)(\alpha d)^{1/2}  \\
&\hspace{-4em}+ \min\left( \frac{ \Psi \Delta_W^2 d }{4m}, \frac{M \Delta_W \sqrt{d}}{2 m} \right)+ A+ \left((1-\alpha m)^K + 1\right)\frac{\Delta_W \sqrt{d}}{2},
\end{align*}
where $A=\sqrt{\frac{(\alpha \Delta_G^2 + \alpha ^{-1} \Delta_W^2) d + 4\alpha \kappa^2}{4m}}$.\endgroup
\end{theorem}
Since the term $A$ contains $\alpha^{-1}$ in the numerator, this theorem implies that as the stepsize $\alpha$ decreases, $W_2$ distance between the stationary distribution of SGLDLP-L and the target distribution may increase. To test if this is the case, we empirically run SGLDLP-L on a standard Gaussian distribution in Figure~\ref{fig:gaussian}. We use 8-bit fixed point and assign 3 of them to represent the fractional part. Our results verify that SGLDLP-L indeed diverges from the target distribution with small stepsizes. In the same time, SGLDLP-F always converges to the target distribution with different stepsizes, aligning with the result in Theorem~\ref{thm:highacc}.

One may choose a stepsize that minimizes the above $W_2$ distance to avoid divergence, however, getting that optimal stepsize is in general difficult since the constants are unknown in practice. Moreover, enabling a small stepsize in SGLD is often desirable, since it is needed to reduce the asymptotic bias of the posterior approximation~\citep{welling2011bayesian}.

\subsection{Variance-Corrected Quantization}
To approach correcting the problem with the na\"ive SGLD with low-precision gradient accumulators, we first need to identify the source of the issue. We show in the following that the reason is the \emph{variance} of each dimension of $\theta_{k+1}$ becomes larger due to using low-precision gradient accumulators.  
Specifically, given the stochastic gradient $\nabla\tilde{U}$, the update of full-precision SGLD can be viewed as sampling from a Gaussian distribution for each dimension $i$
\[
\theta_{k+1,i}\sim\mathcal{N}\left(\theta_{k,i} - \alpha  \nabla\tilde{U}(\theta_{k})_i, 2\alpha \right), \text{for } i=1,\cdots, d.
\]
Since stochastic rounding is unbiased, using it as the weight quantizer $Q_W$ and the gradient quantizer $Q_G$ in SGLDLP-L gives us
  \begingroup\makeatletter\def\f@size{9}\check@mathfonts
\def\maketag@@@#1{\hbox{\m@th\large\normalfont#1}}
\begin{align*}
    \PExv{\theta_{k+1,i}} &= \PExv{Q_W\left(\theta_{k,i} - \alpha Q_G\left(\nabla\tilde{U}(\theta_{k})\right)_i + \sqrt{2\alpha }\xi_{k+1,i}\right)}\\
    &=\theta_{k,i} - \alpha \nabla\tilde{U}(\theta_{k})_i,
\end{align*}\endgroup
which has the same mean as $\theta_{k+1}$ in full-precision. However, the variance of $\theta_{k+1,i}$ is now essentially larger than needed. If we ignore the variance from $Q_G$ and the stochastic gradient, since they are present and have been shown to work well in SGLDLP-F, the variance of $\theta_{k+1,i}$ is
\begingroup\makeatletter\def\f@size{9}\check@mathfonts
\def\maketag@@@#1{\hbox{\m@th\large\normalfont#1}}
\begin{align*}
    &\PVar{\theta_{k+1,i}}\\ 
    &=\PExv{\PVar{Q_W\left(\theta_{k,i} - \alpha \nabla U(\theta_{k})_i + \sqrt{2\alpha }\xi_{k+1,i}\right)\middle|\xi_{k+1,i}}} 
    \\&\hspace{0em}+ \PVar{\PExv{Q_W\left(\theta_{k,i} - \alpha \nabla U(\theta_{k})_i + \sqrt{2\alpha }\xi_{k+1,i}\right)\middle|\xi_{k+1,i}}}\\
    &=\frac{ \Delta_W^2}{4} \chi_{k+1,i} + 2\alpha.
\end{align*}\endgroup
where $\chi_{k+1,i}\in [0,1]$ depends on the distance of $\theta_{k+1,i}$ to its discrete neighbor. The above equation shows that the variance in SGLDLP-L update is larger than the right variance value $2\alpha$. Empirically, from Figure~\ref{fig:gaussian}, we can also find that na\"ive SGLDLP-L estimates the mean correctly but variance wrongly. This validates our intuition that low-precision gradient accumulators with stochastic rounding adds more variance than needed leading to an inaccurate variance estimation.
Besides, we cannot simply use deterministic rounding to solve the problem, since it is a biased estimation and generally provides much worse results than stochastic rounding especially on deep neural networks (see Appendix~\ref{sec:deterministic_rounding_compare} for an empirical demonstration). 

To enable SGLD with low-precision gradient accumulators, we propose a new quantization function $Q^{\text{vc}}$, denoting variance-corrected quantization, to fix the issue. The main idea of $Q^{\text{vc}}$ is to directly sample from the discrete weight space instead of quantizing a real-valued Gaussian sample. First we note that if we want a sample with mean $\mu\ge 0$ and variance $v\le\Delta^2_W /4$, we could sample from the following categorical distribution over $\{\Delta_W,-\Delta_W,0\}$ to get it, 
\begin{align}\label{eq:categorical}
\text{Cat}(\mu,v)=
\begin{cases}
      \Delta_W, &  w.p.  \frac{v+\mu^2+\mu\Delta_W}{2\Delta_W^2}\\
      -\Delta_W, & w.p.  \frac{v+\mu^2-\mu\Delta_W}{2\Delta_W^2} \text{\hspace{1em} }\\ 
      0, & \text{otherwise}
    \end{cases}
\end{align}
Now we show how to use this categorical distribution to preserve the correct mean and variance for quantized $\theta_{k+1}$.
We do so considering two cases: when the Gaussian variance $2\alpha$ is larger than the largest possible stochastic rounding variance $\Delta_W^2/4$, $Q^{vc}$ first adds a small Gaussian noise and uses a sample from Equation~\eqref{eq:categorical} to make up the remaining variance; 
in the other situation, $Q^{vc}$ directly samples from Equation~\eqref{eq:categorical} to achieve the target variance. The full description of $Q^{\text{vc}}$ is outlined in Algorithm~\ref{alg:vc}.

Our variance-corrected quantization function $Q^{\text{vc}}$ always guarantees the correct mean, $\PExv{\theta_{k+1,i}}= \theta_{k,i} - \alpha\nabla\tilde U(\theta_k)_i$, and further guarantees the correct variance $\PVar{\theta_{k+1,i}}= 2\alpha$ most of the time except when $v=2\alpha<v_s$. However that case rarely happens in practice, because the stepsize has to be extremely small. Besides, our quantization is simple to implement and its cost is negligible compared to gradient computation.  Although $Q^{\text{vc}}$ only preserves the correctness of the first two moments (i.e. mean and variance), we show that this does not affect the performance much in both theory and practice. 

\begin{algorithm}[t]
  \caption{Variance-Corrected Low-Precision SGLD (VC SGLDLP-L).}
  \begin{algorithmic}
    \label{alg:vcsgld}
    \STATE \textbf{given:} Stepsize $\alpha$, number of training iterations $K$, gradient quantizer $Q_G$ and quantization gap of weights $\Delta_W$.
    
      \FOR{$k = 1:K$}
	\STATE \textbf{update} $\theta_{k+1} \leftarrow Q^{vc}\left(\theta_{k} - \alpha Q_G\left(\nabla\tilde{U}(\theta_{k})\right), 2\alpha, \Delta_W\right)$ 
    \ENDFOR
    \STATE \textbf{output}: samples $\{\theta_k\}$
    
  \end{algorithmic}
\end{algorithm}

We now prove that SGLDLP-L using $Q^{vc}$, denoting VC SGLDLP-L, converges to the target distribution up to a certain accuracy level with small stepsizes.
\begin{theorem}\label{thm:correction}
We run VC SGLDLP-L as in Algorithm~\ref{alg:vcsgld}. Besides the same assumptions in Theorem~\ref{thm:highacc}, we further assume the gradient is bounded $\mathbf{E}\left[\norm{Q_{G}(\nabla\tilde{U}(\theta_k))}_1\right]\le G$. Let $v_0 = \Delta_W^2/4$. Then
  \begingroup\makeatletter\def\f@size{8}\check@mathfonts
\def\maketag@@@#1{\hbox{\m@th\large\normalfont#1}}
\begin{align*}
    W_2(\mu_K, \pi)&\le (1-\alpha m)^KW_2(\mu_0, \pi) + 1.65 (M/m)(\alpha d)^{1/2} \\&\hspace{-4em}+ \min\left( \frac{\Psi A}{m}, \frac{M\sqrt{A}}{m} \right) 
    + \sqrt{\frac{\alpha \Delta_G^2 d +  4\alpha \kappa^2}{4m} + \frac{A}{\alpha m}} \\&\hspace{-4em}+ \left((1-\alpha m)^K+1\right)\sqrt{A},
\end{align*}
where $A = \begin{cases}
      5 v_0 d, &  \text{if }2\alpha > v_0\\
      \max\left(2\Delta_W\alpha G, 4\alpha d\right), & \text{otherwise}
    \end{cases}$\endgroup
\end{theorem}
This theorem shows that when the stepsize $\alpha \rightarrow 0$, VC SGLDLP-L converges to the target distribution up to an error instead of diverging. Moreover, VC SGLDLP-L converges to the target distribution in $\Ocal(\sqrt{\Delta_W})$ which is equivalent to the convergence rate of SGD with low-precision gradient accumulators to the optimum~\citep{li2017training, yang2019swalp}.  However, we show empirically that VC SGLDLP-L has a much better dependency on the quantization gap than SGD. We leave the improvement of the theoretical bound for future work.

We empirically demonstrate VC SGLDLP-L on the standard Gaussian distribution under the same setting as in the previous section in Figure~\ref{fig:gaussian}. Regardless of the stepsize, VC SGLDLP-L converges to the target distribution and approximates the target distribution as accurately as SGLDLP-F, showing that preserving the correct variance is the key to ensuring correct convergence.

\begin{algorithm}[t]
  \caption{Variance-Corrected Quantization Function $Q^{vc}$.}
  \begin{algorithmic}
  \label{alg:vc}
  \STATE \textbf{input}: ($\mu$, $v$, $\Delta$)  \hspace{0em} \COMMENT{{\color{blue} $Q^{vc}$ returns a variable with mean $\mu$ and variance $v$}}
  \STATE $v_0\leftarrow\Delta^2/4$ \hspace{1em} \COMMENT{{\color{blue} $\Delta^2/4$ is the largest possible variance that stochastic rounding can cause}}
    \IF[{{\color{blue} add a small Gaussian noise and sample from the discrete grid to make up the remaining variance}}]{$v>v_0$}
    \STATE $x \leftarrow \mu + \sqrt{v-v_0}\xi$, where $\xi\sim\mathcal{N}(0,I_d)$
    \STATE $r \leftarrow x - Q^d(x)$ 
    \FORALL{$i$}
    \STATE \textbf{sample} $c_i$ from Cat$(|r_i|,v_0)$ as in Equation~\eqref{eq:categorical}
    \ENDFOR
    \STATE $\theta\leftarrow Q^d(x)+\text{sign}(r)\odot c$  
    \ELSE[{{\color{blue} sample from the discrete grid to achieve the target variance}}]
    \STATE $r \leftarrow \mu - Q^s(\mu)$ 
    \FORALL{$i$}
    \STATE $v_s \leftarrow \left(1-\frac{|r_i|}{\Delta}\right)\cdot r^2_i + \frac{|r_i|}{\Delta}\cdot\left(-r_i+\text{sign}(r_i)\Delta\right)^2$
    \IF{$v>v_s$}
    \STATE \textbf{sample} $c_i$ from Cat$(0,v-v_s)$ as in Equation~\eqref{eq:categorical}
    \STATE $\theta_i\leftarrow  Q^s(\mu)_i + c_i$ 
    \ELSE
    \STATE  $\theta_i\leftarrow Q^s(\mu)_i$
    \ENDIF
    \ENDFOR
    \ENDIF
    \STATE clip $\theta$ if outside representable range
    \STATE \textbf{return} $\theta$
  \end{algorithmic}
\end{algorithm}

\section{Experiments}
We demonstrate the generalization accuracy and uncertainty estimation of low-precision SGLD with full-precision gradient accumulators (SGLDLP-F) and with variance-corrected low-precision gradient accumulators (VC SGLDLP-L) on a logistic regression and multilayer perceptron on MNIST dataset (Section~\ref{sec:mnist}), ResNet-18 on CIFAR datasets and LSTM on IMDB dataset (Section~\ref{sec:cifar}), and ResNet-18 on ImageNet dataset (Section~\ref{sec:imagenet}). We use \texttt{qtorch}~\citep{zhang2019qpytorch} to simulate low-precision training on these experiments, and use the same quantization for weights, gradients, activations, backpropagation errors unless otherwise stated. For all experiments, SGLD collects samples from the posterior of the model’s weight, and obtained the prediction on test data by Bayesian model averaging. We mainly compare low-precision SGLD with low-precision SGD which has been used to achieve state-of-the-art results in low-precision deep learning~\citep{sun2019hybrid,sun2020ultra}. We use SGLDFP and SGDFP to denote SGLD and SGD in full-precision respectively.

\begin{figure}[t]
	\vspace{-0mm}\centering
	\begin{tabular}{cccc}		
		\includegraphics[width=6cm]{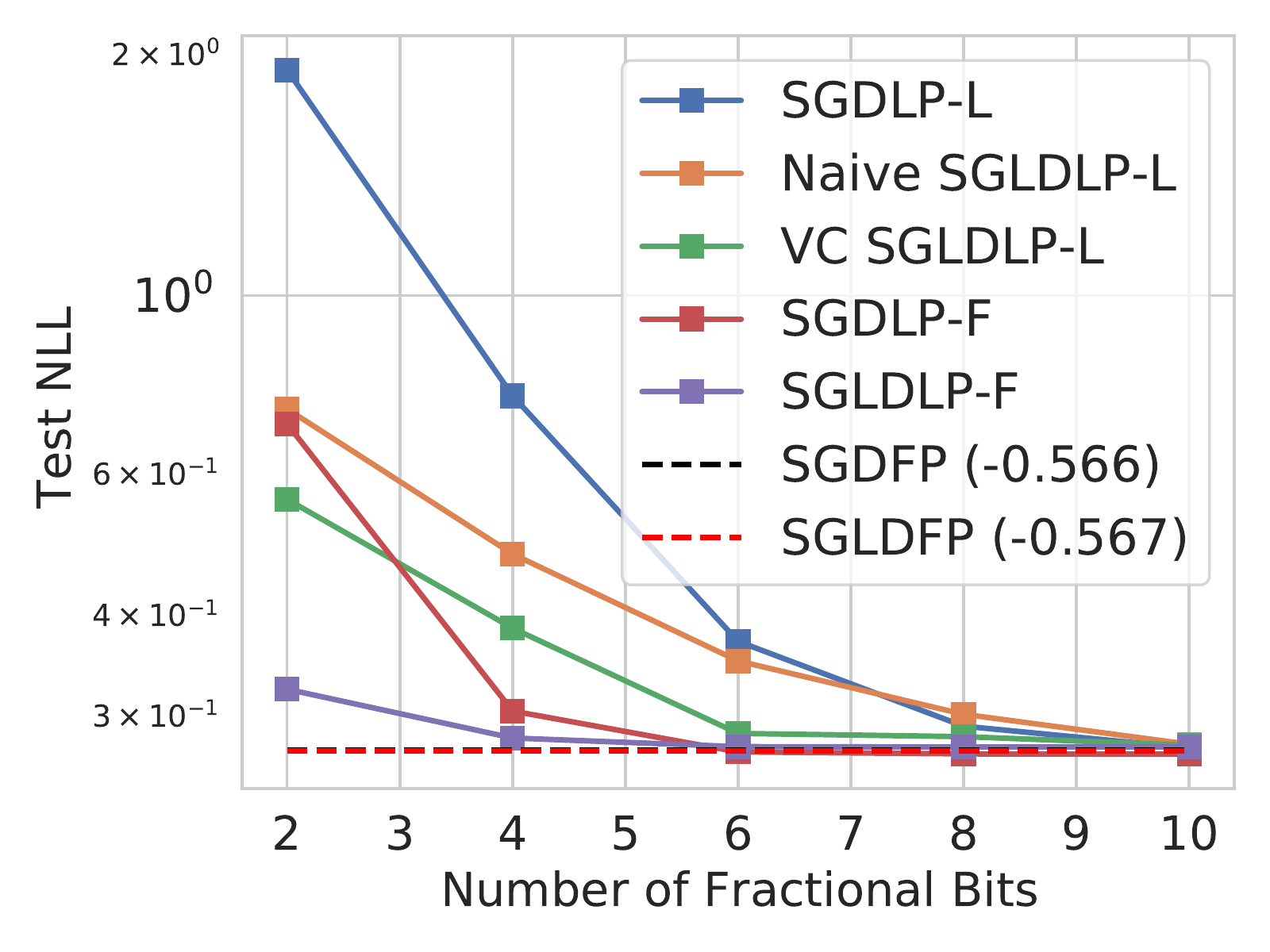}   \\
		(a) Logistic regression \\
		\includegraphics[width=6cm]{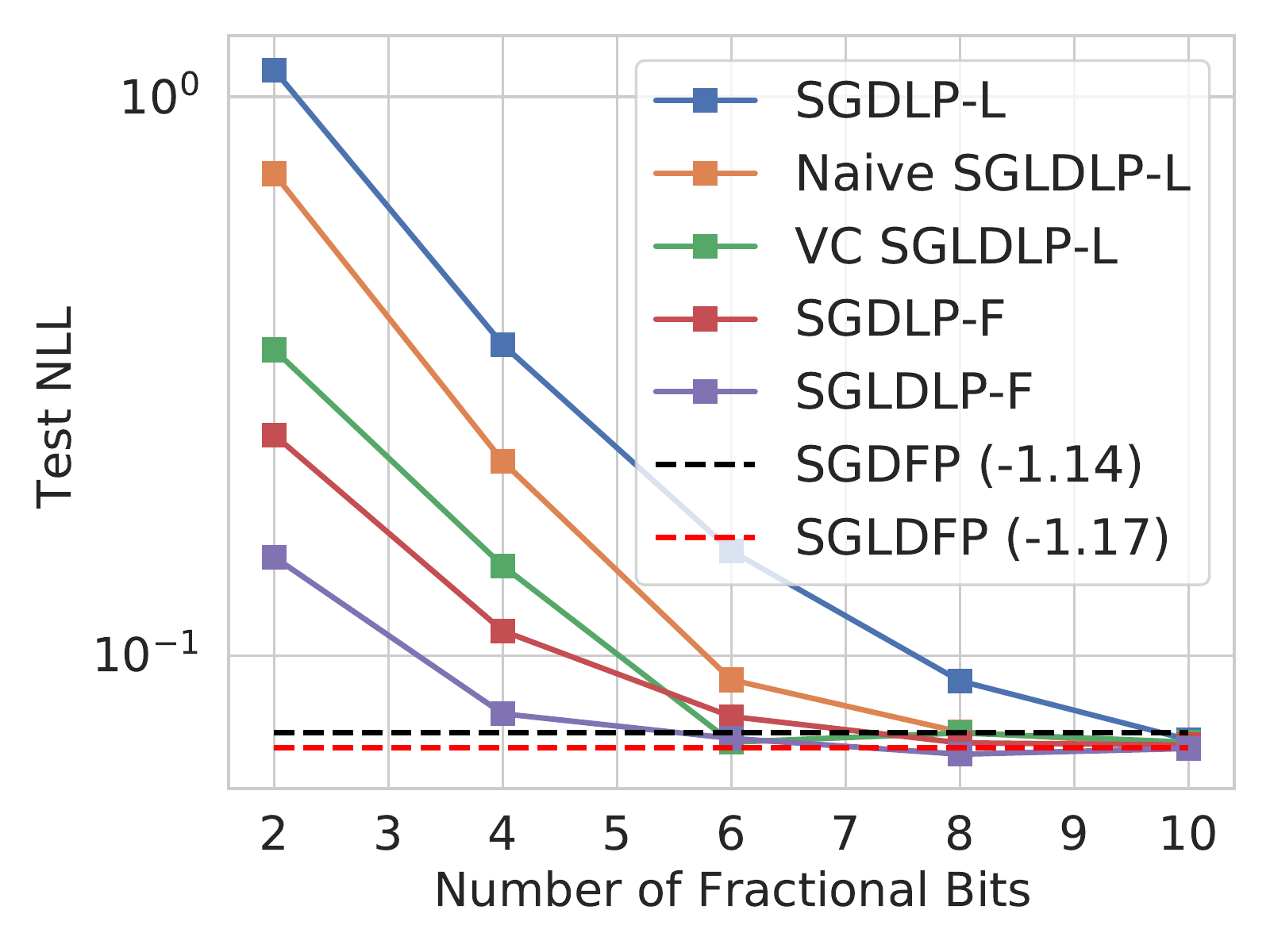}      \\
		
		(b) MLP\\
	\end{tabular}
	\caption{Test NLL on MNIST in terms of different precision. As the quantization level increases, VC SGLDLP-L and SGLDLP-F are able to maintain high performance whereas the corresponding SGD deteriorates quickly.}
	\label{fig:mnist}
	\vspace{-1em}
\end{figure}
\subsection{MNIST}\label{sec:mnist}

\paragraph{Logistic Regression} We first empirically verify our theorems, including the bias dependence on the quantization levels and the tightness of the bounds, on a logistic regression on MNIST dataset. We use $\mathcal{N}(0,1/6)$ as the prior distribution following~\citet{yang2019swalp} and the resulting posterior distribution is strongly log-concave and satisfies the assumptions in Section~\ref{sec:sgldlp}. We use fixed point numbers with 2 integer bits and vary the number of fractional bits which is corresponding to varying the quantization gap $\Delta$. We report test negative log-likelihood (NLL) with different numbers of fractional bits in Figure~\ref{fig:mnist}a.

From the results, we see that SGLDLP-F is more robust to the decay of the number of bits than its SGD counterpart since SGLDLP-F outperforms SGDLP-F on all number of bits and recovers the full-precision result with 6 bits whereas SGDLP-F needs 10 bits. This verifies Theorem~\ref{thm:highacc} that SGLDLP-F converges to the target distribution up to an error and is more robust to the quantization gap than SGDLP-F. This also shows that the bound in Theorem~\ref{thm:highacc} is relatively tight in terms of the quantization gap since SGLDLP-F needs roughly half number of bits compared to SGDLP-F to achieve the same convergence accuracy. With low-precision gradient accumulators, we can see that VC SGLDLP-L is significantly better than na\"ive SGLDLP-L, indicating that our variance-corrected quantization effectively reduces the bias of gradient accumulators, which verifies Theorem~\ref{thm:lowacc} and Theorem~\ref{thm:correction}. Moreover, VC SGLDLP-L outperforms SGDLP-L on all numbers of bits and even outperforms SGDLP-F when using 2 fractional bits. These observations demonstrate that with either full- or low-precision gradient accumulators, SGLD is able to maintain its high performance whereas SGD deteriorates quickly as the quantization noise increases.

\paragraph{Multilayer Perceptron}
To test whether these results apply to non-log-concave distributions, we replace the logistic regression model by a two-layer multilayer perceptron (MLP). The MLP has 100 hidden units and RELU nonlinearities. From Figure~\ref{fig:mnist}b, we observe similar results as on the logistic regression, suggesting that empirically our analysis still stands on general distributions and sheds light on the possibility of extending the theoretical analysis of low-precision SGLD to non-log-concave settings. Please note that as far as we understand, there is no theoretical convergence result of low-precision optimization in non-convex settings either.

\begin{table*}[t]
\begin{minipage}[t]{0.55 \linewidth}
\caption{Test errors (\%) on CIFAR with ResNet-18 and IMDB with LSTM. Low-precision SGLD outperforms low-precision SGD across different datasets, architectures and number representations, and the improvement becomes larger when using more low-precision arithmetic.}
\label{tab:cifar}
\vskip 0.15in
\begin{center}
\begin{sc}
\resizebox{0.98\linewidth}{!}{
\begin{tabular}{lccccc}
\toprule
  &CIFAR-10 &CIFAR-100 &IMDB \\
\midrule
32-bit Floating Point\\
\midrule
SGLDFP &4.65 {\scriptsize$\pm$0.06}  &22.58 {\scriptsize$\pm$0.18}   &13.43 {\scriptsize$\pm$0.21}\\
SGDFP  &4.71 {\scriptsize$\pm$0.02}  &22.64 {\scriptsize$\pm$0.13}   &13.88 {\scriptsize$\pm$0.29}\\
\hdashline
cSGLDFP &4.54 {\scriptsize$\pm$0.05} &21.63 {\scriptsize$\pm$0.04} &13.25 {\scriptsize$\pm$0.18}  \\
\toprule
8-bit Fixed Point\\
\midrule
Na\"ive SGLDLP-L &7.82 {\scriptsize$\pm$0.13}  &27.25 {\scriptsize$\pm$0.13}  &16.63 {\scriptsize$\pm$0.28}\\
VC SGLDLP-L &7.13 {\scriptsize$\pm$0.01} &26.62 {\scriptsize$\pm$0.16}   &15.38 {\scriptsize$\pm$0.27}\\
SGDLP-L &8.53 {\scriptsize$\pm$0.08}    &28.86 {\scriptsize$\pm$	0.10}   &19.28 {\scriptsize$\pm$0.63}\\
SGLDLP-F &5.12 {\scriptsize$\pm$0.06} &23.30 {\scriptsize$\pm$0.09}   &15.40 {\scriptsize$\pm$0.36}\\
SGDLP-F &5.20 {\scriptsize$\pm$0.14}  &23.84 {\scriptsize$\pm$0.12}   &15.74 {\scriptsize$\pm$0.79}\\
\toprule
8-bit Block Floating Point\\
\midrule
Na\"ive SGLDLP-L &5.85 {\scriptsize$\pm$0.04}  &26.38 {\scriptsize$\pm$0.13} &14.64 {\scriptsize$\pm$0.08} \\
VC SGLDLP-L &5.51 {\scriptsize$\pm$0.01} &25.22 {\scriptsize$\pm$0.18}   &13.99 {\scriptsize$\pm$0.24}\\
SGDLP-L &5.86 {\scriptsize$\pm$0.18}   &26.19 {\scriptsize$\pm$0.11} &16.06 {\scriptsize$\pm$1.81}\\
SGLDLP-F &4.58 {\scriptsize$\pm$0.07} &22.59 {\scriptsize$\pm$0.18} &14.05 {\scriptsize$\pm$0.33}\\
SGDLP-F &4.75 {\scriptsize$\pm$0.05}  &22.9 {\scriptsize$\pm$0.13} &14.28 {\scriptsize$\pm$0.17}\\
\hdashline
VC cSGLDLP-L &4.97 {\scriptsize$\pm$0.10} &22.61 {\scriptsize$\pm$0.12} & 13.09 {\scriptsize$\pm$0.27}\\
cSGLD-F &4.32 {\scriptsize$\pm$0.07}  &21.50  {\scriptsize$\pm$0.14} &13.13 {\scriptsize$\pm$0.37}\\
\bottomrule
\end{tabular}
}
\end{sc}
\end{center}
  
  \end{minipage}
 \hfill
 \begin{minipage}[t]{0.44 \linewidth}
 \caption{ECE $\downarrow$ (\%) on CIFAR with ResNet-18. VC SGLDLP-L and SGLDLP-F achieve almost the same or even lower ECE than full-precision SGLD whereas the ECE of low-precision SGD increases significantly.}
\label{tab:ece}
\vskip 0.15in
\begin{center}
\begin{small}
\begin{sc}
\resizebox{0.98\linewidth}{!}{
\begin{tabular}{lccc|ccc}
\toprule
  &CIFAR-10 & CIFAR-100\\
\midrule
32-bit Floating Point\\
\midrule
SGLD &1.11 &3.92\\
SGD & 2.53 &4.97 \\
\hdashline
cSGLDFP &0.66   &1.38\\
\toprule
8-bit Fixed Point\\
\midrule
VC SGLDLP-L &0.6 &3.19\\
SGDLP-L &3.4 & 10.38\\
SGLDLP-F &1.12 &4.42\\
SGDLP-F &3.05 & 6.80\\
\toprule
8-bit Block Floating Point\\
\midrule
VC SGLDLP-L &0.6 &5.82\\
SGDLP-L &4.23 & 12.97\\
SGLDLP-F &1.19 &3.78\\
SGDLP-F &2.76 & 5.2\\
\hdashline
VC cSGLDLP-L &0.51 &1.39\\
cSGLD-F &0.56   &1.33\\
\bottomrule
\end{tabular}
}
\end{sc}
\end{small}
\end{center}

 \end{minipage}
 
\end{table*}

\subsection{CIFAR and IMDB}\label{sec:cifar}
We consider image and sentiment classification tasks: CIFAR datasets~\citep{krizhevsky2009learning} on ResNet-18~\citep{he2016deep}, and IMDB dataset~\citep{maas2011learning} on LSTM~\citep{hochreiter1997long}. We use 8-bit number representation since it becomes increasingly popular in training deep models and is powered by the new generation of chips~\citep{sun2019hybrid,banner2018scalable,wang2018training}. We report test errors averaged over 3 runs with the standard error in Table~\ref{tab:cifar}. 

\paragraph{Fixed Point} We use 8-bit fixed point for weights and gradients but full-precision for activations since we find low-precision activations significantly harm the performance. Similar to the results in previous sections, SGLDLP-F is better than SGDLP-F and VC SGLDLP-L significantly outperforms na\"ive SGLDLP-L and SGDLP-L across datasets and architectures. Notably, the improvement of SGLD over SGD becomes larger when using more low-precision arithmetic. For example, on CIFAR-100, VC SGLDLP-L outperforms SGDLP-L by 2.24\%, SGLDLP-F outperforms SGDLP-F by 0.54\% and SGLDFP outperforms SGDFP by 0.06\%. This demonstrates that  SGLD is particularly compatible with low-precision deep learning because of its natural ability to handle system noise.

\paragraph{Block Floating Point} 
We also consider block floating point (BFP) which is another common number type and is often preferred over fixed point on deep models due to less quantization error caused by overflow and underflow ~\citep{song2018computation}. Following the block design in~\citet{yang2019swalp}, we use \emph{small-block} for ResNet and \emph{big-block} for LSTM. The $Q^{vc}$ function naturally generalizes to BFP and only needs a small modification (see Appendix~\ref{sec:bfp} for the algorithm of $Q^{vc}$ with BFP). By using BFP, the results of all low-precision methods improve over fixed point. SGLDLP-F can match the performance of SGLDFP with all numbers quantized to 8-bit except gradient accumulators. VC SGLDLP-L still outperforms na\"ive SGLDLP-L indicating the effectiveness of $Q^{vc}$ with BFP. Again, SGLDFP-F and VC SGLDLP-L outperform their SGD counterparts on all tasks, suggesting the general applicability of low-precision SGLD with different number types.

\paragraph{Cyclical SGLD} We further apply low-precision to a recent variant of SGLD, \emph{cSGLD}, which utilizes a cyclical learning rate schedule to speed up convergence~\citep{zhang2019cyclical}. We observe that the results of cSGLD-F are very close to those of cSGLDFP, and VC cSGLDLP-L can match or even outperforms full-precision SGD with all numbers quantized to 8 bits! These results indicate that diverse samples from different modes, obtained by the cyclical learning rate schedule, can counter-effect the quantization error by providing complementary predictions. 

\paragraph{Expected Calibration Error}
Besides generalization performance, we further report the results of expected calibration error (ECE)~\citep{guo2017calibration} to demonstrate the uncertainty estimation of low-precision SGLD. In Table~\ref{tab:ece}, we observe that SGLDLP-F and VC SGLDLP-L achieve almost the same or even lower ECE than full-precision SGLD, showing the ability of SGLD to give well-calibrated predictions does not degenerate due to using low-precision. VC SGLDLP-L sometimes gives lower ECE than SGLDLP-F which may be due to the regularization effect of low-precision arithmetic. Moreover, cSGLD in low-precision not only achieves the best accuracy but also has the best calibration, further suggesting that diverse samples obtained by a cyclical learning rate schedule have a positive effect on quantization. In contrast, the ECE of low-precision SGD increases significantly compared to full-precision SGD, implying that the quantization error makes the standard DNNs even more overconfident, which might lead to wrong decisions in real-world applications.

\paragraph{SGLDLP-F vs VC SGLDLP-L}
We have provided two variants of low-precision SGLD for practical use. In general, SGLDLP-F has better performance while VC SGLDLP-L requires less computation, making them suitable for different cases. When the computation resources are very limited, e.g. on edge devices, VC SGLDLP-L is preferred for saving computation while when the resources are able to support full-precision gradient accumulators, SGLDLP-F is preferred for better performance.

\subsection{ImageNet}\label{sec:imagenet}
Finally, we test low-precision SGLD on a large-scale image classification dataset, ImageNet, with ResNet-18. We train SGD for 90 epochs and train SGLD for 10 epochs using the trained SGD model as the initialization. In Table~\ref{tab:imagenet}, we observe that the improvement of SGLD over SGD is larger in low-precision (0.76\% top-1 error) than in full-precision (0.17\% top-1 error), showing the advantages of low-precision SGLD on large-scale deep learning tasks. We could not achieve reasonable results with low-precision gradient accumulators for SGD and SGLD, which might be caused by hyper-parameter tuning.

\begin{table}[t]

\vskip -0.1in
\end{table}

\begin{table}[t]
\caption{Test errors (\%) on ImageNet with ResNet-18. The improvement of SGLD over SGD becomes larger in low-precision than in full-precision. }
\label{tab:imagenet}
\vskip 0.15in
\begin{center}
\begin{small}
\begin{sc}
\begin{tabular}{lccc|ccc}
\toprule
  &Top-1 & Top-5\\
\midrule
32-bit Floating Point\\
\midrule
SGLD &30.39 &10.76\\
SGD & 30.56 &10.97 \\
\toprule
8-bit Block Floating Point\\
\midrule
SGLDLP-F &31.47 &11.77\\
SGDLP-F &32.23 & 12.09\\
\bottomrule
\end{tabular}
\end{sc}
\end{small}
\end{center}
\vskip -0.1in
\end{table}
\section{Conclusion}

We provide the first comprehensive investigation for low-precision SGLD. With full-precision gradient accumulators, we prove that SGLD is convergent and can be safely used in practice, and further show that it has a better dependency of convergence on the quantization gap than SGD. 
Moreover, we reveal issues in na\"ively performing low-precision computation in SGLD with low-precision gradient accumulators, and propose a new theoretically guaranteed quantization function to enable fully quantized sampling.
We conduct experiments on a Gaussian distribution and a logistic regression to empirically verify our theoretical results. Besides, we show that low-precision SGLD achieves comparable results with full-precision SGLD and outperforms low-precision SGD significantly on several Bayesian deep learning benchmarks.

MCMC was once the gold standard on small neural networks~\citep{neal2011mcmc}, but has been significantly limited by its high costs on large architectures in deep learning. We believe this work fills an important gap, and will accelerate the practical use of sampling methods on large-scale and resource-restricted machine learning problems. 

Moreover, low-precision SGLD could broadly be used as a drop-in replacement for standard SGLD, as it can confer speed and memory advantages, while retaining accuracy.

\section*{Acknowledgements}
RZ is supported by NSF AI Institute for Foundations of Machine Learning (IFML). AGW is supported by NSF CAREER IIS-2145492, NSF I-DISRE 193471, NIH R01DA048764-01A1, NSF IIS-1910266, Meta Core Data Science, Google AI Research, BigHat Biosciences, Capital One, and an Amazon Research Award.

\bibliography{example_paper}

\begin{thebibliography}{52}
\providecommand{\natexlab}[1]{#1}
\providecommand{\url}[1]{\texttt{#1}}
\expandafter\ifx\csname urlstyle\endcsname\relax
  \providecommand{\doi}[1]{doi: #1}\else
  \providecommand{\doi}{doi: \begingroup \urlstyle{rm}\Url}\fi

\bibitem[Achterhold et~al.(2018)Achterhold, Koehler, Schmeink, and
  Genewein]{achterhold2018variational}
Achterhold, J., Koehler, J.~M., Schmeink, A., and Genewein, T.
\newblock Variational network quantization.
\newblock In \emph{International Conference on Learning Representations}, 2018.

\bibitem[Ahn et~al.(2014)Ahn, Shahbaba, and Welling]{ahn2014distributed}
Ahn, S., Shahbaba, B., and Welling, M.
\newblock Distributed stochastic gradient mcmc.
\newblock In \emph{International conference on machine learning}, pp.\
  1044--1052. PMLR, 2014.

\bibitem[Baker et~al.(2019)Baker, Fearnhead, Fox, and Nemeth]{baker2019control}
Baker, J., Fearnhead, P., Fox, E.~B., and Nemeth, C.
\newblock Control variates for stochastic gradient mcmc.
\newblock \emph{Statistics and Computing}, 29\penalty0 (3):\penalty0 599--615,
  2019.

\bibitem[Banner et~al.(2018)Banner, Hubara, Hoffer, and
  Soudry]{banner2018scalable}
Banner, R., Hubara, I., Hoffer, E., and Soudry, D.
\newblock Scalable methods for 8-bit training of neural networks.
\newblock \emph{Advances in neural information processing systems}, 2018.

\bibitem[Cai et~al.(2018)Cai, Ren, Liu, Ding, Wang, Qian, Pedram, and
  Wang]{cai2018vibnn}
Cai, R., Ren, A., Liu, N., Ding, C., Wang, L., Qian, X., Pedram, M., and Wang,
  Y.
\newblock Vibnn: Hardware acceleration of bayesian neural networks.
\newblock \emph{ACM SIGPLAN Notices}, 53\penalty0 (2):\penalty0 476--488, 2018.

\bibitem[Chen et~al.(2016)Chen, Ding, Li, Zhang, and Carin]{chen2016stochastic}
Chen, C., Ding, N., Li, C., Zhang, Y., and Carin, L.
\newblock Stochastic gradient mcmc with stale gradients.
\newblock \emph{Advances in Neural Information Processing Systems}, 2016.

\bibitem[Chen et~al.(2014)Chen, Fox, and Guestrin]{chen2014stochastic}
Chen, T., Fox, E., and Guestrin, C.
\newblock Stochastic gradient hamiltonian monte carlo.
\newblock In \emph{International conference on machine learning}, pp.\
  1683--1691. PMLR, 2014.

\bibitem[Cheng et~al.(2015)Cheng, Soudry, Mao, and Lan]{cheng2015training}
Cheng, Z., Soudry, D., Mao, Z., and Lan, Z.
\newblock Training binary multilayer neural networks for image classification
  using expectation backpropagation.
\newblock \emph{arXiv preprint arXiv:1503.03562}, 2015.

\bibitem[Courbariaux et~al.(2015)Courbariaux, Bengio, and
  David]{courbariaux2015binaryconnect}
Courbariaux, M., Bengio, Y., and David, J.-P.
\newblock Binaryconnect: Training deep neural networks with binary weights
  during propagations.
\newblock In \emph{Advances in neural information processing systems}, pp.\
  3123--3131, 2015.

\bibitem[Dalalyan \& Karagulyan(2019)Dalalyan and Karagulyan]{dalalyan2019user}
Dalalyan, A.~S. and Karagulyan, A.
\newblock User-friendly guarantees for the langevin monte carlo with inaccurate
  gradient.
\newblock \emph{Stochastic Processes and their Applications}, 129\penalty0
  (12):\penalty0 5278--5311, 2019.

\bibitem[De~Sa et~al.(2017)De~Sa, Feldman, R{\'e}, and
  Olukotun]{de2017understanding}
De~Sa, C., Feldman, M., R{\'e}, C., and Olukotun, K.
\newblock Understanding and optimizing asynchronous low-precision stochastic
  gradient descent.
\newblock In \emph{Proceedings of the 44th Annual International Symposium on
  Computer Architecture}, pp.\  561--574, 2017.

\bibitem[Deng et~al.(2020)Deng, Lin, and Liang]{deng2020contour}
Deng, W., Lin, G., and Liang, F.
\newblock A contour stochastic gradient langevin dynamics algorithm for
  simulations of multi-modal distributions.
\newblock \emph{Advances in neural information processing systems},
  33:\penalty0 15725--15736, 2020.

\bibitem[Dubey et~al.(2016)Dubey, J~Reddi, Williamson, Poczos, Smola, and
  Xing]{dubey2016variance}
Dubey, K.~A., J~Reddi, S., Williamson, S.~A., Poczos, B., Smola, A.~J., and
  Xing, E.~P.
\newblock Variance reduction in stochastic gradient langevin dynamics.
\newblock \emph{Advances in neural information processing systems},
  29:\penalty0 1154--1162, 2016.

\bibitem[Esser et~al.(2020)Esser, McKinstry, Bablani, Appuswamy, and
  Modha]{esser2019learned}
Esser, S.~K., McKinstry, J.~L., Bablani, D., Appuswamy, R., and Modha, D.~S.
\newblock Learned step size quantization.
\newblock \emph{International Conference on Learning Representations}, 2020.

\bibitem[Ferianc et~al.(2021)Ferianc, Maji, Mattina, and
  Rodrigues]{ferianc2021effects}
Ferianc, M., Maji, P., Mattina, M., and Rodrigues, M.
\newblock On the effects of quantisation on model uncertainty in bayesian
  neural networks.
\newblock \emph{Uncertainty in Artificial Intelligence}, 2021.

\bibitem[Gan et~al.(2017)Gan, Li, Chen, Pu, Su, and Carin]{gan2016scalable}
Gan, Z., Li, C., Chen, C., Pu, Y., Su, Q., and Carin, L.
\newblock Scalable bayesian learning of recurrent neural networks for language
  modeling.
\newblock \emph{Association for Computational Linguistics}, 2017.

\bibitem[Guo et~al.(2017)Guo, Pleiss, Sun, and Weinberger]{guo2017calibration}
Guo, C., Pleiss, G., Sun, Y., and Weinberger, K.~Q.
\newblock On calibration of modern neural networks.
\newblock In \emph{International Conference on Machine Learning}, pp.\
  1321--1330. PMLR, 2017.

\bibitem[Gupta et~al.(2015)Gupta, Agrawal, Gopalakrishnan, and
  Narayanan]{gupta2015deep}
Gupta, S., Agrawal, A., Gopalakrishnan, K., and Narayanan, P.
\newblock Deep learning with limited numerical precision.
\newblock In \emph{International conference on machine learning}, pp.\
  1737--1746. PMLR, 2015.

\bibitem[He et~al.(2016)He, Zhang, Ren, and Sun]{he2016deep}
He, K., Zhang, X., Ren, S., and Sun, J.
\newblock Deep residual learning for image recognition.
\newblock In \emph{Proceedings of the IEEE conference on computer vision and
  pattern recognition}, pp.\  770--778, 2016.

\bibitem[Heek \& Kalchbrenner(2019)Heek and Kalchbrenner]{heek2019bayesian}
Heek, J. and Kalchbrenner, N.
\newblock Bayesian inference for large scale image classification.
\newblock \emph{arXiv preprint arXiv:1908.03491}, 2019.

\bibitem[Hochreiter \& Schmidhuber(1997)Hochreiter and
  Schmidhuber]{hochreiter1997long}
Hochreiter, S. and Schmidhuber, J.
\newblock Long short-term memory.
\newblock \emph{Neural computation}, 9\penalty0 (8):\penalty0 1735--1780, 1997.

\bibitem[Jacob et~al.(2018)Jacob, Kligys, Chen, Zhu, Tang, Howard, Adam, and
  Kalenichenko]{jacob2018quantization}
Jacob, B., Kligys, S., Chen, B., Zhu, M., Tang, M., Howard, A., Adam, H., and
  Kalenichenko, D.
\newblock Quantization and training of neural networks for efficient
  integer-arithmetic-only inference.
\newblock In \emph{Proceedings of the IEEE conference on computer vision and
  pattern recognition}, pp.\  2704--2713, 2018.

\bibitem[Korattikara et~al.(2015)Korattikara, Rathod, Murphy, and
  Welling]{korattikara2015bayesian}
Korattikara, A., Rathod, V., Murphy, K., and Welling, M.
\newblock Bayesian dark knowledge.
\newblock \emph{arXiv preprint arXiv:1506.04416}, 2015.

\bibitem[Krishnamoorthi(2018)]{krishnamoorthi2018quantizing}
Krishnamoorthi, R.
\newblock Quantizing deep convolutional networks for efficient inference: A
  whitepaper.
\newblock \emph{arXiv preprint arXiv:1806.08342}, 2018.

\bibitem[Krizhevsky et~al.(2009)Krizhevsky, Hinton,
  et~al.]{krizhevsky2009learning}
Krizhevsky, A., Hinton, G., et~al.
\newblock Learning multiple layers of features from tiny images.
\newblock 2009.

\bibitem[Li et~al.(2016)Li, Stevens, Chen, Pu, Gan, and Carin]{li2016learning}
Li, C., Stevens, A., Chen, C., Pu, Y., Gan, Z., and Carin, L.
\newblock Learning weight uncertainty with stochastic gradient mcmc for shape
  classification.
\newblock In \emph{Proceedings of the IEEE Conference on Computer Vision and
  Pattern Recognition}, pp.\  5666--5675, 2016.

\bibitem[Li et~al.(2019)Li, Chen, Pu, Henao, and Carin]{li2019communication}
Li, C., Chen, C., Pu, Y., Henao, R., and Carin, L.
\newblock Communication-efficient stochastic gradient mcmc for neural networks.
\newblock In \emph{Proceedings of the AAAI Conference on Artificial
  Intelligence}, volume~33, pp.\  4173--4180, 2019.

\bibitem[Li et~al.(2017)Li, De, Xu, Studer, Samet, and
  Goldstein]{li2017training}
Li, H., De, S., Xu, Z., Studer, C., Samet, H., and Goldstein, T.
\newblock Training quantized nets: A deeper understanding.
\newblock \emph{Advances in neural information processing systems}, 2017.

\bibitem[Li \& De~Sa(2019)Li and De~Sa]{li2019dimension}
Li, Z. and De~Sa, C.~M.
\newblock Dimension-free bounds for low-precision training.
\newblock \emph{Advances in Neural Information Processing Systems}, 2019.

\bibitem[Lin et~al.(2016)Lin, Talathi, and Annapureddy]{lin2016fixed}
Lin, D., Talathi, S., and Annapureddy, S.
\newblock Fixed point quantization of deep convolutional networks.
\newblock In \emph{International conference on machine learning}, pp.\
  2849--2858. PMLR, 2016.

\bibitem[Ma et~al.(2015)Ma, Chen, and Fox]{ma2015complete}
Ma, Y.-A., Chen, T., and Fox, E.~B.
\newblock A complete recipe for stochastic gradient mcmc.
\newblock \emph{Advances in neural information processing systems}, 2015.

\bibitem[Ma et~al.(2019)Ma, Chen, Jin, Flammarion, and Jordan]{ma2019sampling}
Ma, Y.-A., Chen, Y., Jin, C., Flammarion, N., and Jordan, M.~I.
\newblock Sampling can be faster than optimization.
\newblock \emph{Proceedings of the National Academy of Sciences}, 116\penalty0
  (42):\penalty0 20881--20885, 2019.

\bibitem[Maas et~al.(2011)Maas, Daly, Pham, Huang, Ng, and
  Potts]{maas2011learning}
Maas, A., Daly, R.~E., Pham, P.~T., Huang, D., Ng, A.~Y., and Potts, C.
\newblock Learning word vectors for sentiment analysis.
\newblock In \emph{Proceedings of the 49th annual meeting of the association
  for computational linguistics: Human language technologies}, pp.\  142--150,
  2011.

\bibitem[Meng et~al.(2020)Meng, Bachmann, and Khan]{meng2020training}
Meng, X., Bachmann, R., and Khan, M.~E.
\newblock Training binary neural networks using the bayesian learning rule.
\newblock In \emph{International Conference on Machine Learning}, pp.\
  6852--6861. PMLR, 2020.

\bibitem[Micikevicius et~al.(2018)Micikevicius, Narang, Alben, Diamos, Elsen,
  Garcia, Ginsburg, Houston, Kuchaiev, Venkatesh,
  et~al.]{micikevicius2017mixed}
Micikevicius, P., Narang, S., Alben, J., Diamos, G., Elsen, E., Garcia, D.,
  Ginsburg, B., Houston, M., Kuchaiev, O., Venkatesh, G., et~al.
\newblock Mixed precision training.
\newblock \emph{International Conference on Learning Representations}, 2018.

\bibitem[Neal et~al.(2011)]{neal2011mcmc}
Neal, R.~M. et~al.
\newblock Mcmc using hamiltonian dynamics.
\newblock \emph{Handbook of markov chain monte carlo}, 2\penalty0
  (11):\penalty0 2, 2011.

\bibitem[Song et~al.(2018)Song, Liu, and Wang]{song2018computation}
Song, Z., Liu, Z., and Wang, D.
\newblock Computation error analysis of block floating point arithmetic
  oriented convolution neural network accelerator design.
\newblock In \emph{Proceedings of the AAAI Conference on Artificial
  Intelligence}, volume~32, 2018.

\bibitem[Soudry et~al.(2014)Soudry, Hubara, and Meir]{soudry2014expectation}
Soudry, D., Hubara, I., and Meir, R.
\newblock Expectation backpropagation: Parameter-free training of multilayer
  neural networks with continuous or discrete weights.
\newblock In \emph{Advances in Neural Information Processing Systems},
  volume~1, pp.\ ~2, 2014.

\bibitem[Su et~al.(2019)Su, Cvitkovic, and Huang]{su2019sampling}
Su, J., Cvitkovic, M., and Huang, F.
\newblock Sampling-free learning of bayesian quantized neural networks.
\newblock \emph{arXiv preprint arXiv:1912.02992}, 2019.

\bibitem[Sun et~al.(2019)Sun, Choi, Chen, Wang, Venkataramani, Srinivasan, Cui,
  Zhang, and Gopalakrishnan]{sun2019hybrid}
Sun, X., Choi, J., Chen, C.-Y., Wang, N., Venkataramani, S., Srinivasan, V.~V.,
  Cui, X., Zhang, W., and Gopalakrishnan, K.
\newblock Hybrid 8-bit floating point (hfp8) training and inference for deep
  neural networks.
\newblock \emph{Advances in neural information processing systems},
  32:\penalty0 4900--4909, 2019.

\bibitem[Sun et~al.(2020)Sun, Wang, Chen, Ni, Agrawal, Cui, Venkataramani,
  El~Maghraoui, Srinivasan, and Gopalakrishnan]{sun2020ultra}
Sun, X., Wang, N., Chen, C.-Y., Ni, J., Agrawal, A., Cui, X., Venkataramani,
  S., El~Maghraoui, K., Srinivasan, V.~V., and Gopalakrishnan, K.
\newblock Ultra-low precision 4-bit training of deep neural networks.
\newblock \emph{Advances in Neural Information Processing Systems}, 33, 2020.

\bibitem[Talwar(2019)]{talwar2019computational}
Talwar, K.
\newblock Computational separations between sampling and optimization.
\newblock \emph{Advances in neural information processing systems}, 2019.

\bibitem[van Baalen et~al.(2020)van Baalen, Louizos, Nagel, Amjad, Wang,
  Blankevoort, and Welling]{van2020bayesian}
van Baalen, M., Louizos, C., Nagel, M., Amjad, R.~A., Wang, Y., Blankevoort,
  T., and Welling, M.
\newblock Bayesian bits: Unifying quantization and pruning.
\newblock \emph{Advances in Neural Information Processing Systems}, 2020.

\bibitem[Wang et~al.(2018{\natexlab{a}})Wang, Vicol, Lucas, Gu, Grosse, and
  Zemel]{wang2018adversarial}
Wang, K.-C., Vicol, P., Lucas, J., Gu, L., Grosse, R., and Zemel, R.
\newblock Adversarial distillation of bayesian neural network posteriors.
\newblock In \emph{International Conference on Machine Learning}, pp.\
  5190--5199. PMLR, 2018{\natexlab{a}}.

\bibitem[Wang et~al.(2018{\natexlab{b}})Wang, Choi, Brand, Chen, and
  Gopalakrishnan]{wang2018training}
Wang, N., Choi, J., Brand, D., Chen, C.-Y., and Gopalakrishnan, K.
\newblock Training deep neural networks with 8-bit floating point numbers.
\newblock In \emph{Proceedings of the 32nd International Conference on Neural
  Information Processing Systems}, pp.\  7686--7695, 2018{\natexlab{b}}.

\bibitem[Welling \& Teh(2011)Welling and Teh]{welling2011bayesian}
Welling, M. and Teh, Y.~W.
\newblock Bayesian learning via stochastic gradient langevin dynamics.
\newblock In \emph{Proceedings of the 28th international conference on machine
  learning (ICML-11)}, pp.\  681--688. Citeseer, 2011.

\bibitem[Wu et~al.(2018)Wu, Li, Chen, and Shi]{wu2018training}
Wu, S., Li, G., Chen, F., and Shi, L.
\newblock Training and inference with integers in deep neural networks.
\newblock \emph{International Conference on Learning Representations}, 2018.

\bibitem[Yang et~al.(2019)Yang, Zhang, Kirichenko, Bai, Wilson, and
  De~Sa]{yang2019swalp}
Yang, G., Zhang, T., Kirichenko, P., Bai, J., Wilson, A.~G., and De~Sa, C.
\newblock Swalp: Stochastic weight averaging in low-precision training.
\newblock \emph{International Conference on Machine Learning}, 2019.

\bibitem[Zhang et~al.(2020)Zhang, Li, Zhang, Chen, and
  Wilson]{zhang2019cyclical}
Zhang, R., Li, C., Zhang, J., Chen, C., and Wilson, A.~G.
\newblock Cyclical stochastic gradient mcmc for bayesian deep learning.
\newblock \emph{International Conference on Learning Representations}, 2020.

\bibitem[Zhang et~al.(2019)Zhang, Lin, Yang, and De~Sa]{zhang2019qpytorch}
Zhang, T., Lin, Z., Yang, G., and De~Sa, C.
\newblock Qpytorch: A low-precision arithmetic simulation framework.
\newblock \emph{arXiv preprint arXiv:1910.04540}, 2019.

\bibitem[Zhou et~al.(2016)Zhou, Wu, Ni, Zhou, Wen, and Zou]{zhou2016dorefa}
Zhou, S., Wu, Y., Ni, Z., Zhou, X., Wen, H., and Zou, Y.
\newblock Dorefa-net: Training low bitwidth convolutional neural networks with
  low bitwidth gradients.
\newblock \emph{arXiv preprint arXiv:1606.06160}, 2016.

\bibitem[Zhu et~al.(2019)Zhu, Dong, and Su]{zhu2019binary}
Zhu, S., Dong, X., and Su, H.
\newblock Binary ensemble neural network: More bits per network or more
  networks per bit?
\newblock In \emph{Proceedings of the IEEE/CVF Conference on Computer Vision
  and Pattern Recognition}, pp.\  4923--4932, 2019.

\end{thebibliography}
\bibliographystyle{icml2022}

\newpage
\appendix
\onecolumn

\section{Quantization Formulation}\label{sec:lp-formulation}
We follow the quantization framework in prior work~\citep{wu2018training,wang2018training,yang2019swalp} to quantize weights, activations, backpropagation errors, and gradients, as outlined in Algorithm~\ref{alg:lp-framework}.

\begin{algorithm}[H]
  \caption{Low-Precision Training for SGLD.}
  \begin{algorithmic}
    \label{alg:lp-framework}
    \STATE \textbf{given:} $L$ layers DNN $\{f_1\ldots,f_L\}$.
    Stepsize $\alpha$. Weight, gradient, activation and error quantizers $Q_W,Q_G,Q_A,Q_E$. Variance-corrected quantization $Q^{vc}$, deterministic rounding $Q^d$, stochastic rounding $Q^s$ and quantization gap of weights $\Delta_W$. Data batch sequence $\{(x_k,y_k)\}_{k=1}^K$. $\theta_{k}^{fp}$ denotes the full-precision buffer of the weight.

      \FOR{$k = 1:K$}
      \STATE \textbf{1. Forward Propagation:}
	\STATE \hspace{2em} $a_k^{(0)}=x_k$
	\STATE\hspace{2em} $a_k^{(l)}=Q_{A}(f_l(a_k^{(l-1)},\theta_k^{l})), \forall l\in [1,L]$
	\STATE \textbf{2. Backward Propagation:} 
	\STATE \hspace{2em} $e^{(L)}=\nabla_{a_k^{(L)}}\Lcal(a_k^{(L)},y_k)$
	\STATE\hspace{2em} $e^{(l-1)}=Q_{E}\left(\frac{\partial f_l(a_k^{(l)})}{\partial a_k^{(l-1)} }e_k^{(l)}\right), \forall l\in [1,L]$
	\STATE\hspace{2em} $g_k^{(l)}=Q_G\left(\frac{\partial f_l}{\partial \theta_k^{(l)} }e_k^{(l)}\right), \forall l\in [1,L]$
	\STATE \textbf{3. SGLD Update:}
	\STATE\hspace{2em} \textbf{full-precision gradient accumulators: }$\theta_{k+1}^{fp} \leftarrow \theta_{k}^{fp} - \alpha Q_G\left(\nabla\tilde{U}(\theta_{k})\right)+ \sqrt{2\alpha}\xi$,\hspace{1em} $\theta_{k+1} \leftarrow Q_{W}\left(\theta_{k+1}^{fp}\right)$
	\STATE\hspace{2em} \textbf{low-precision gradient accumulators: }$\theta_{k+1} \leftarrow Q^{vc}\left(\theta_{k} - \alpha Q_G\left(\nabla\tilde{U}(\theta_{k})\right), 2\alpha, \Delta_W\right)$ 
    \ENDFOR
    \STATE \textbf{output}: samples $\{\theta_k\}$
    
  \end{algorithmic}
\end{algorithm}

\section{Proof of Theorem~\ref{thm:highacc}}\label{sec:proof_thm1}
Our proofs in the paper follow Theorem 4 in~\citet{dalalyan2019user}, which provides a convergence bound of Langevin dynamics with noisy gradients. We state the result of Theorem 4 in~\citet{dalalyan2019user} below.

We consider Langevin dynamics whose update rule is
\begin{align}\label{eq:dalalyan}
    \theta_{k+1} = \theta_k - \alpha \left(\nabla U(\theta_k) + \zeta_k\right) + \sqrt{2\alpha}\xi_{k+1}.
\end{align}

The noise in the gradient $\zeta_k$ has the following  three assumptions:
\begin{align*}
    \PExv{\norm{\PExv{\zeta_k\middle|\theta_k}}_2^2}\le \delta^2 d, \hspace{1em} \PExv{ \norm{\zeta_k -\PExv{\zeta_k\middle|\theta_k}}_2^2}\le \sigma^2 d, \hspace{1em} \xi_{k+1} \text{ is independent of } (\zeta_0, \cdots, \zeta_k),
\end{align*}
where $\delta > 0$ and $\sigma>0$ are some constants. Under the same assumptions in Section~\ref{sec:sgldlp}, we have the convergence bound for the above Langevin dynamics.
\begin{theorem}[Theorem 4 in \citet{dalalyan2019user}]\label{thm:dalalyan}
We run the above Langevin dynamics with $\alpha\le 2/(m+M)$. Let $\pi$ be the target distribution, $\mu_0$ be the initial distribution and $\mu_K$ be the distribution obtained by the Langevin dynamics in Equation~\eqref{eq:dalalyan} after $K$ iterations. Then
\begin{align*}
    W_2(\mu_K, \pi) &\le (1-\alpha m)^KW_2(\mu_0, \pi) + 1.65 (M/m)(\alpha d)^{1/2} + \frac{\delta\sqrt{d}}{m} + \frac{\sigma^2 (\alpha d)^{1/2}}{1.65M + \sigma\sqrt{m}}.
\end{align*}
\end{theorem}

Buit upon this theorem, we now prove Theorem~\ref{thm:highacc}.

\begin{proof}
We write the SGLDLP-F update as the following
\begin{align*}
    \theta_{k+1} &= \theta_k - \alpha Q_{G}(\nabla\tilde{U}(Q_{W}(\theta_k))) + \sqrt{2\alpha }\xi_{k+1}\\
&=\theta_k - \alpha \left(\nabla U(\theta_k) + \zeta_k\right) + \sqrt{2\alpha }\xi_{k+1}
\end{align*}

where
\begin{align*}
    \zeta_k &= Q_{G}(\nabla\tilde{U}(Q_{W}(\theta_k))) - \nabla U(\theta_k)\\
    &=
    Q_{G}(\nabla\tilde{U}(Q_{W}(\theta_k))) - \nabla\tilde{U}(Q_{W}(\theta_k)) 
    \\&\hspace{2em}+ \nabla\tilde{U}(Q_{W}(\theta_k)) - \nabla U(Q_{W}(\theta_k)) 
    + \nabla U(Q_{W}(\theta_k)) - \nabla U(\theta_k).
\end{align*}

Since $\mathbf{E}[\nabla\tilde{U}(x)] = \nabla U(x)$ and $\mathbf{E}[Q(x)] = x$, we have
\begin{align*}
    \PExv{\zeta_k\middle|\theta_k} 
    &=
   \PExv{Q_{G}(\nabla\tilde{U}(Q_{W}(\theta_k))) - \nabla\tilde{U}(Q_{W}(\theta_k))\middle|\theta_k}
     \\&\hspace{2em}+ \PExv{\nabla\tilde{U}(Q_{W}(\theta_k)) - \nabla U(Q_{W}(\theta_k))\middle|\theta_k} + \PExv{\nabla U(Q_{W}(\theta_k)) - \nabla U(\theta_k)\middle|\theta_k}\\
    &=\PExv{\nabla U(Q_{W}(\theta_k)) - \nabla U(\theta_k)\middle|\theta_k}.
\end{align*}

By the assumption, we know that
\begin{align*}
    \norm{\nabla U(Q_{W}(\theta_k)) - \nabla U(\theta_k) }^2_2 &\le M^2 \norm{Q_{W}(\theta_k) - \theta_k}^2_2,
\end{align*}

then it follows that
\begin{align*}
    \norm{\PExv{\zeta_k\middle|\theta_k}}_2^2
    &=
    \norm{\PExv{\nabla U(Q_{W}(\theta_k)) - \nabla U(\theta_k)\middle|\theta_k}}_2^2\\
    & \le \PExv{\norm{\nabla U(Q_{W}(\theta_k)) - \nabla U(\theta_k) }^2_2\middle|\theta_k} \\
    &\le M^2 \PExv{\norm{Q_{W}(\theta_k) - \theta_k}^2_2\middle|\theta_k}\\
    &\le M^2 \cdot \frac{\Delta_W^2d}{4}.
\end{align*}

Let $f: \R \rightarrow \R^d$ denote the function
\[
    f(a) = \nabla U(\theta_k + a(Q_{W}(\theta_k) - \theta_k)).
\]
By the mean value theorem, there will exist an $a \in [0,1]$ (a function of the weight quantization randomness) such that
\[
    f(1) - f(0) = f'(a).
\]
So,
\begin{align*}
    \PExv{\zeta_k\middle|\theta_k}
    &=
    \PExv{\nabla U(Q_{W}(\theta_k)) - \nabla U(\theta_k)\middle|\theta_k} \\
    &=
    \PExv{ \nabla^2 U(\theta_k + a(Q_{W}(\theta_k) - \theta_k)) (Q_{W}(\theta_k) - \theta_k) \middle|\theta_k}\\
    &=
    \PExv{ \nabla^2 U(\theta_k) (Q_{W}(\theta_k) - \theta_k) \middle|\theta_k}
    \\&\hspace{2em}+
    \PExv{ \left(
        \nabla^2 U(\theta_k + a(Q_{W}(\theta_k) - \theta_k))
        -
        \nabla^2 U(\theta_k)
    \right)
    (Q_{W}(\theta_k) - \theta_k) \middle|\theta_k}\\
    &=
    \PExv{ \left(
        \nabla^2 U(\theta_k + a(Q_{W}(\theta_k) - \theta_k))
        -
        \nabla^2 U(\theta_k)
    \right)
    (Q_{W}(\theta_k) - \theta_k) \middle|\theta_k}.
\end{align*}

Now, by the assumption $\| \nabla^2 U(x) - \nabla^2 U(y) \|_2 \le \Psi \| x - y \|_2$, we get

\begin{align*}
    \norm{\PExv{\zeta_k\middle|\theta_k}}_2
    &=
    \norm{\PExv{ \left(
        \nabla^2 U(\theta_k + a(Q_{W}(\theta_k) - \theta_k))
        -
        \nabla^2 U(\theta_k)
    \right)
    (Q_{W}(\theta_k) - \theta_k) \middle|\theta_k}}_2
    \\&\le
    \PExv{\norm{\left(
        \nabla^2 U(\theta_k + a(Q_{W}(\theta_k) - \theta_k))
        -
        \nabla^2 U(\theta_k)
    \right)
    (Q_{W}(\theta_k) - \theta_k)}_2 \middle|\theta_k}
    \\&\le
    \PExv{ \norm{
        \nabla^2 U(\theta_k + a(Q_{W}(\theta_k) - \theta_k))
        -
        \nabla^2 U(\theta_k)
    }_2
    \norm{ Q_{W}(\theta_k) - \theta_k)}_2 \middle|\theta_k}
    \\&\le
    \PExv{ \Psi \norm{ a(Q_{W}(\theta_k) - \theta_k) }_2
    \norm{ Q_{W}(\theta_k) - \theta_k)}_2 \middle|\theta_k}
    \\&\le
    \Psi \PExv{
    \norm{ Q_{W}(\theta_k) - \theta_k}_2^2 \middle|\theta_k}
    \\&\le
    \frac{ \Psi \Delta_W^2 d }{4}.
\end{align*}

This combined with the previous result gives us
\[
    \norm{\PExv{\zeta_k\middle|\theta_k}}_2 \le \min\left( \frac{ \Psi \Delta_W^2 d }{4}, \frac{M \Delta_W \sqrt{d}}{2} \right).
\]

Now considering the variance of $\zeta_k$,
\begin{align*}
    &\hspace{-2em}\mathbf{E}\left[\norm{\zeta_k - \PExv{\zeta_k\middle|\theta_k} }_2^2\right]\\
    &\le
    \mathbf{E}\left[\norm{\zeta_k }_2^2\right] \\
    &\le 
    \mathbf{E}\left[\norm{Q_{G}(\nabla\tilde{U}(Q_{W}(\theta_k))) - \nabla\tilde{U}(Q_{W}(\theta_k)) }_2^2\right]
    \\&\hspace{2em}+
    \mathbf{E}\left[\norm{\nabla\tilde{U}(Q_{W}(\theta_k)) - \nabla U(Q_{W}(\theta_k))  }_2^2\right]
    +
    \mathbf{E}\left[\norm{\nabla U(Q_{W}(\theta_k)) - \nabla U(\theta_k)  }_2^2\right]\\
    & \le
    \frac{\Delta_G^2 d}{4} + \kappa^2 + M^2 \cdot \frac{\Delta_W^2 d}{4}.
\end{align*}

Recall that to apply the result of Theorem 4 in \citet{dalalyan2019user}, we need
\begin{align*}
    \PExv{\norm{\PExv{\zeta_k\middle|\theta_k}}_2^2}\le 
    \delta^2 d, \hspace{1em} \PExv{ \norm{\zeta_k -\PExv{\zeta_k\middle|\theta_k}}_2^2}\le \sigma^2 d.
\end{align*}

We set $\delta$ and $\sigma$ to be
\[
\delta = \min\left( \frac{ \Psi \Delta_W^2 \sqrt{d} }{4}, \frac{M \Delta_W}{2} \right), \hspace{1em} \sigma^2d = \frac{\Delta_G^2 d}{4} + \kappa^2 + M^2 \cdot \frac{\Delta_W^2 d}{4} = \frac{(\Delta_G^2 + M^2 \Delta_W^2) d + 4\kappa^2}{4}.
\]
Since $\zeta_k$ is independent of the Gaussian noise $\xi_{i}, \text{ for }i=0,\ldots,k+1$, we have shown that the assumptions in Theorem 4 in \citet{dalalyan2019user} are satisfied. Thus we apply the result in Theorem 4 and get  
\begin{align*}
    W_2(\mu_K, \pi) &\le (1-\alpha m)^KW_2(\mu_0, \pi) + 1.65 (M/m)(\alpha d)^{1/2} + \frac{\delta\sqrt{d}}{m} + \frac{\sigma^2 (\alpha d)^{1/2}}{1.65M + \sigma\sqrt{m}}\\
    &\le (1-\alpha m)^KW_2(\mu_0, \pi) + 1.65 (M/m)(\alpha d)^{1/2} + \frac{\delta\sqrt{d}}{m} + \sqrt{\frac{\sigma^2 \alpha d}{m}}\\
    &=
    (1-\alpha m)^KW_2(\mu_0, \pi) + 1.65 (M/m)(\alpha d)^{1/2} + \min\left( \frac{ \Psi \Delta_W^2 d }{4m}, \frac{M \Delta_W \sqrt{d}}{2 m} \right) \\
    &\hspace{2em}+ \sqrt{\frac{(\Delta_G^2 + M^2 \Delta_W^2) \alpha d + 4\alpha \kappa^2}{4m}}.
\end{align*}
Note that we ignore the $1.65M$ in the denominator to further simplify the bound. 
\end{proof}

\section{Proof of Theorem~\ref{thm:lowacc}}\label{sec:proof_thm2}
\begin{proof}
Recall that the update of SGLDLP-L is
\begin{align*}
    \theta_{k+1} &= Q_{W}\left(\theta_k - \alpha  Q_{G}(\nabla\tilde{U}(\theta_k)) + \sqrt{2\alpha }\xi_{k+1}\right).
\end{align*}
To utilize the result in \citet{dalalyan2019user}, we introduce an intermediate dynamic $\psi_{k+1} = \theta_k - \alpha Q_{G}(\nabla\tilde{U}(\theta_k)) + \sqrt{2\alpha }\xi_{k+1}$. Therefore $\theta_k = Q_{W}(\psi_k)$ and
\begin{align*}
    \psi_{k+1} &= \theta_k - \alpha Q_{G}(\nabla\tilde{U}(\theta_k)) + \sqrt{2\alpha }\xi_{k+1}\\
    &= 
    Q_{W}(\psi_k) - \alpha Q_{G}(\nabla\tilde{U}(Q_{W}(\psi_k))) + \sqrt{2\alpha }\xi_{k+1}\\
    &= 
    \psi_k - \alpha (\nabla U(\psi_k) + \zeta_k)  + \sqrt{2\alpha }\xi_{k+1}
\end{align*}
where
\begin{align*}
    \zeta_k 
    &=
    \frac{\psi_k - \theta_k}{\alpha } + Q_{G}(\nabla\tilde{U}(\theta_k)) - \nabla U(\psi_k)
    \\&=
    \frac{\psi_k - \theta_k}{\alpha } + Q_{G}(\nabla\tilde{U}(\theta_k)) - \nabla\tilde{U}(\theta_k) 
    + \nabla\tilde{U}(\theta_k) - \nabla U(\theta_k) 
    + \nabla U(\theta_k) - \nabla U(\psi_k).
\end{align*}

Similar to the previous proof in Section~\ref{sec:proof_thm1}, we know that 
\[
    \PExv{\zeta_k\middle|\psi_k}
    =\PExv{\nabla U(\theta_k) - \nabla U(\psi_k)\middle|\psi_k}
    =\PExv{\nabla U(Q_{W}(\psi_k)) - \nabla U(\psi_k)\middle|\psi_k},
\]
so
\begin{align*}
    \norm{\PExv{\zeta_k\middle|\psi_k} }_2
    &\le \min\left( \frac{\Psi \Delta_W^2 d}{4}, \frac{M \Delta_W \sqrt{d}}{2} \right),
\end{align*}
and it suffices to set $\delta$ the same as in Section~\ref{sec:proof_thm1}.
On the other hand, the variance will be bounded by
\begin{align*}
    \mathbf{E}\left[\norm{\zeta_k - \PExv{\zeta_k \middle|\psi_k}}_2^2\right]
    &\le
    \mathbf{E}\left[ \norm{ Q_{G}(\nabla\tilde{U}(\theta_k)) - \nabla\tilde{U}(\theta_k) }_2^2 \right]
    +
    \mathbf{E}\left[ \norm{ \nabla\tilde{U}(\theta_k) - \nabla U(\theta_k)  }_2^2 \right]
    \\&\hspace{2em}+
    \mathbf{E}\left[ \norm{ \frac{\psi_k - \theta_k}{\alpha } 
    + \nabla U(\theta_k) - \nabla U(\psi_k) }_2^2 \right]
    \\&\le
    \frac{\Delta_G^2 d}{4} + \kappa^2 + 
    \mathbf{E}\left[ \norm{ \nabla F(\psi_k) - \nabla F(\theta_k) }_2^2  \right],
\end{align*}
where $F(\theta) = \frac{1}{2\alpha } \norm{\theta}_2^2 - U(\theta)$.
Observe that since $U$ is $m$-strongly convex and $M$-smooth, and $\alpha ^{-1} \ge M/2$, $F$ must be $\alpha^{-1}$-smooth, and so
\begin{align*}
    \mathbf{E}\left[\norm{\zeta_k - \PExv{\zeta_k \middle|\psi_k} }_2^2\right]
    &\le
    \frac{\Delta_G^2 d}{4} + \kappa^2 + 
    \frac{1}{\alpha^2} \mathbf{E}\left[ \norm{ \psi_k - \theta_k }_2^2  \right] \\
    &\le
    \frac{\Delta_G^2 d}{4} + \kappa^2 + 
    \frac{\Delta_W^2 d}{4 \alpha^2}.
\end{align*}

This is essentially replacing the $M^2$ in the previous analysis with $\alpha^{-2}$. It suffices to set $\sigma^2 d= \frac{\Delta_G^2 d}{4} + \kappa^2 + 
    \frac{\Delta_W^2 d}{4 \alpha^2}$.

Supposing the distribution of $\psi_{K+1}$ is $\nu_K$, applying Theorem 4 in \citet{dalalyan2019user} will give us the rate of
\begin{align*}
    W_2(\nu_K, \pi) &\le
    (1-\alpha m)^KW_2(\nu_0, \pi) + 1.65 (M/m)(\alpha d)^{1/2} + \min\left( \frac{ \Psi \Delta_W^2 d }{4m}, \frac{M \Delta_W \sqrt{d}}{2 m} \right) \\
    &\hspace{2em}+ \sqrt{\frac{(\alpha \Delta_G^2 + \alpha^{-1} \Delta_W^2) d + 4\alpha \kappa^2}{4m}}.
\end{align*}

We also have
\begin{align*}
    W_2(\mu_k, \nu_k) = \left(\inf_{J\in \mathcal{J}(x,y)}\int \norm{x-y}_2^2dJ(x,y)\right)^{1/2} \le \mathbf{E}\left[ \norm{ \theta_{k+1} - \psi_{k+1} }_2^2 \right]^{\frac{1}{2}} \le \frac{\Delta_W \sqrt{d}}{2}.
\end{align*}
Combining these two results, we get the final bound
\begin{align*}
    W_2(\mu_K, \pi) &\le W_2(\mu_K, \nu_K) + W_2(\nu_K, \pi) \\
    &\le
    (1-\alpha m)^KW_2(\nu_0, \pi) + 1.65 (M/m)(\alpha d)^{1/2} + \min\left( \frac{ \Psi \Delta_W^2 d }{4m}, \frac{M \Delta_W \sqrt{d}}{2 m} \right)  \\
    &+ \sqrt{\frac{(\alpha \Delta_G^2 + \alpha^{-1} \Delta_W^2) d + 4\alpha \kappa^2}{4m}}+ \frac{\Delta_W \sqrt{d}}{2}\\
    &\le
    (1-\alpha m)^KW_2(\mu_0, \pi) + 1.65 (M/m)(\alpha d)^{1/2} + \min\left( \frac{ \Psi \Delta_W^2 d }{4m}, \frac{M \Delta_W \sqrt{d}}{2 m} \right)  \\
    &+ \sqrt{\frac{(\alpha \Delta_G^2 + \alpha^{-1} \Delta_W^2) d + 4\alpha \kappa^2}{4m}}+ \left((1-\alpha m)^K + 1\right)\frac{\Delta_W \sqrt{d}}{2}.
\end{align*}
\end{proof}

\section{Proof of Theorem~\ref{thm:correction}}\label{sec:proof_thm3}
\begin{proof}
Recall that the update of VC SGLDLP-L is
\begin{align*}
    \theta_{k+1} &= Q^{vc}\left(\theta_k - \alpha  Q_{G}(\nabla\tilde{U}(\theta_k)), 2\alpha, \Delta_W\right).
\end{align*}
We ignore the variance of $Q_G$ since it is relatively small compared to the weight quantization variance in practice. $Q^{vc}$ is defined as in Algorithm~\ref{alg:vc} and we have $\PExv{\theta_{k+1}\middle|\theta_k} = \theta_k - \alpha  \nabla U(\theta_k)$.

Let $\psi_{k+1} = \theta_k - \alpha Q_{G}(\nabla\tilde{U}(\theta_k)) + \sqrt{2\alpha }\xi_{k+1}$ then it follows that
\[
\psi_{k+1}-\theta_{k+1}= \theta_k - \alpha Q_{G}(\nabla\tilde{U}(\theta_k)) + \sqrt{2\alpha }\xi_{k+1} - \theta_{k+1},
\]
and
\begin{align*}
    \psi_{k+1}
    &= 
    \psi_k - \alpha (\nabla U(\psi_k) + \zeta_k)  + \sqrt{2\alpha }\xi_{k+1}
\end{align*}
where
\begin{align*}
    \zeta_k 
    &=
    \frac{\psi_k - \theta_k}{\alpha } + Q_{G}(\nabla\tilde{U}(\theta_k)) - \nabla U(\psi_k)
    \\&=
    \frac{\psi_k - \theta_k}{\alpha } + Q_{G}(\nabla\tilde{U}(\theta_k)) - \nabla\tilde{U}(\theta_k) 
    + \nabla\tilde{U}(\theta_k) - \nabla U(\theta_k) 
    + \nabla U(\theta_k) - \nabla U(\psi_k).
\end{align*}

Note that $\mathbf{E}[\psi_k-\theta_k]=0$. Similar to the previous proof in Section~\ref{sec:proof_thm1}, we know that
\begin{align*}
    \norm{\PExv{\zeta_k\middle|\psi_k} }_2^2
    =\norm{\PExv{\nabla U(\theta_k) - \nabla U(\psi_k)\middle|\psi_k}}_2^2\le M^2\PExv{\norm{\psi_k - \theta_k}^2_2\middle|\psi_k}.
\end{align*}

When $2\alpha > v_0 = \frac{\Delta_W^2}{4}$, we have that
\begin{align*}
    &\PExv{\norm{\psi_k-\theta_k}_2^2\middle|\psi_k} \\
    &= \PExv{\norm{\left(\theta_{k-1} - \alpha  Q_{G}(\nabla\tilde{U}(\theta_{k-1}))\right) + \sqrt{2\alpha}\xi_k - Q^d\left(\theta_{k-1} - \alpha  Q_{G}(\nabla\tilde{U}(\theta_{k-1})) + \sqrt{2\alpha - v_0}\xi_k\right) - \text{sign}(r)c}_2^2\middle|\psi_k}.
\end{align*}
Let 
\begin{align*}
b &= Q^d\left(\theta_{k-1} - \alpha  Q_{G}(\nabla\tilde{U}(\theta_{k-1})) + \sqrt{2\alpha - v_0}\xi_k\right)
\\&\hspace{2em}-\left(\theta_{k-1} - \alpha  Q_{G}(\nabla\tilde{U}(\theta_{k-1})) + \sqrt{2\alpha - v_0}\xi_k\right),
\end{align*}
then $\Abs{b}\le \frac{\Delta_W}{2}$ and
\begin{align*}
    &\PExv{\norm{\psi_k-\theta_k}_2^2\middle|\psi_k} \\
    &= \PExv{\norm{\left(\theta_{k-1} - \alpha  Q_{G}(\nabla\tilde{U}(\theta_{k-1}))\right) + \sqrt{2\alpha}\xi_k - \left(\theta_{k-1} - \alpha  Q_{G}(\nabla\tilde{U}(\theta_{k-1})) + \sqrt{2\alpha - v_0}\xi_k\right) -b - \text{sign}(r)c}_2^2\middle|\psi_k}\\
    &=\PExv{\norm{ \sqrt{2\alpha}\xi_k - \sqrt{2\alpha - v_0}\xi_k - b- \text{sign}(r)c}_2^2\middle|\psi_k}\\
    &\le \PExv{\norm{ \sqrt{2\alpha}\xi_k - \sqrt{2\alpha - v_0}\xi_k - b}_2^2\middle|\psi_k}+ \PExv{\norm{\text{sign}(r)c}_2^2\middle|\psi_k}\\
    &\le \mathbf{E}\left[\norm{ \Abs{\sqrt{2\alpha}\xi_k - \sqrt{2\alpha - v_0}\xi_k }+ \frac{\Delta_W}{2}}_2^2\right] + v_0d\\
    &\le (\sqrt{2\alpha} - \sqrt{2\alpha - v_0})^2\mathbf{E}[\norm{\xi_k}_2^2] + (\sqrt{2\alpha} - \sqrt{2\alpha - v_0})\Delta_W \mathbf{E}[\norm{\xi_k}_2] + 2v_0d\\
    &\le \left((\sqrt{2\alpha} - \sqrt{2\alpha - v_0})^2 + (\sqrt{2\alpha} - \sqrt{2\alpha - v_0})\Delta_W + 2v_0 \right)d.
\end{align*}
Since $2xy\le x^2+y^2$, we get
\[
(\sqrt{2\alpha} - \sqrt{2\alpha - v_0})\Delta_W \le (\sqrt{2\alpha} - \sqrt{2\alpha - v_0})^2 + \frac{\Delta_W^2}{4} = (\sqrt{2\alpha} - \sqrt{2\alpha - v_0})^2 + v_0.
\]
The expression can be further simplified to be
\begin{align*}
    \PExv{\norm{\psi_k-\theta_k}_2^2\middle|\psi_k}
    &\le \left(2(\sqrt{2\alpha} - \sqrt{2\alpha - v_0})^2 + 3v_0 \right)d.
\end{align*}
We also note that
\[
\sqrt{2\alpha} - \sqrt{2\alpha - v_0} = \frac{2\alpha - (2\alpha - v_0)}{\sqrt{2\alpha} + \sqrt{2\alpha - v_0}} = \frac{v_0}{\sqrt{2\alpha} + \sqrt{2\alpha - v_0}}\le \frac{v_0}{\sqrt{2\alpha}},
\]
then the expectation becomes
\begin{align*}
    \PExv{\norm{\psi_k-\theta_k}_2^2\middle|\psi_k}
    &\le \left(\frac{v_0^2}{\alpha} + 3v_0 \right)d.
\end{align*}
Since $2\alpha > v_0$, it follows that
\begin{align*}
    \PExv{\norm{\psi_k-\theta_k}_2^2\middle|\theta_k}
    &\le \left(2v_0 + 3v_0 \right)d = 5 v_0 d.
\end{align*}

Let $A = 5 v_0 d$. Then we obtain
\begin{align*}
    \norm{\PExv{\zeta_k\middle|\psi_k} }_2^2
    \le M^2 \cdot A,
\end{align*}

and
\begin{align*}
    \norm{\PExv{\zeta_k\middle|\psi_k} }_2
    \le \Psi \cdot A.
\end{align*}
Therefore, it suffices to set 
\[
    \delta = \min\left( \Psi A, M\sqrt{A} \right).
\]
We now consider the variance which will be bounded by
\begin{align*}
    \mathbf{E}\left[\norm{\zeta_k - \PExv{\zeta_k \middle|\psi_k} }_2^2\right]
    &\le
    \mathbf{E}\left[ \norm{ Q_{G}(\nabla\tilde{U}(\theta_k)) - \nabla\tilde{U}(\theta_k) }_2^2 \right]
    +
    \mathbf{E}\left[ \norm{ \nabla\tilde{U}(\theta_k) - \nabla U(\theta_k)  }_2^2 \right]
    \\&\hspace{2em}+
    \mathbf{E}\left[ \norm{ \frac{\psi_k - \theta_k}{\alpha } 
    + \nabla U(\theta_k) - \nabla U(\psi_k) }_2^2 \right]
    \\&\le
    \frac{\Delta_G^2 d}{4} + \kappa^2 + 
    \frac{1}{\alpha^2} \mathbf{E}\left[ \norm{ \psi_k - \theta_k }_2^2  \right]
    \\&\le
    \frac{\Delta_G^2 d}{4} + \kappa^2 + 
    \frac{A}{\alpha^2}.
\end{align*}
It suffices to set $\sigma^2 d= \frac{\Delta_G^2 d}{4} + \kappa^2 + 
    \frac{A}{\alpha^2}$.
Supposing the distribution of $\psi_{K+1}$ is $\nu_K$, applying Theorem 4 in \citet{dalalyan2019user} will give us the rate of
\begin{align*}
    W_2(\nu_K, \pi) &\le
    (1-\alpha m)^KW_2(\nu_0, \pi) + 1.65 (M/m)(\alpha d)^{1/2} + \min\left( \frac{\Psi \cdot A}{m}, \frac{M\sqrt{A}}{m} \right) \\
    &\hspace{2em}+ \sqrt{\frac{\alpha \Delta_G^2 d +  4\alpha \kappa^2}{4m} + \frac{A}{\alpha m}}.
\end{align*}
We also have
\begin{align*}
    W_2(\mu_K, \nu_K) = \left(\inf_{J\in \mathcal{J}(x,y)}\int \norm{x-y}_2^2dJ(x,y)\right)^{1/2} \le \mathbf{E}\left[ \norm{ \theta_{K+1} - \psi_{K+1} }_2^2 \right]^{\frac{1}{2}} \le \sqrt{A}.
\end{align*}
Combining these two results, we get
\begin{align*}
    W_2(\mu_K, \pi) &\le W_2(\mu_K, \nu_K) + W_2(\nu_K, \pi) \\
    &\le
    (1-\alpha m)^KW_2(\nu_0, \pi) + 1.65 (M/m)(\alpha d)^{1/2} + \min\left( \frac{\Psi \cdot A}{m}, \frac{M\sqrt{A}}{m} \right) \\
    &\hspace{2em}+ \sqrt{\frac{\alpha \Delta_G^2 d +  4\alpha \kappa^2}{4m} + \frac{A}{\alpha m}} + \sqrt{A}\\
    &\le
    (1-\alpha m)^KW_2(\mu_0, \pi) + 1.65 (M/m)(\alpha d)^{1/2} + \min\left( \frac{\Psi \cdot A}{m}, \frac{M\sqrt{A}}{m} \right) \\
    &\hspace{2em}+ \sqrt{\frac{\alpha \Delta_G^2 d +  4\alpha \kappa^2}{4m} + \frac{A}{\alpha m}} + \left((1-\alpha m)^K+1\right)\sqrt{A}.
\end{align*}

When $2\alpha < \frac{\Delta_W^2}{4}$, since we assume that the gradient is bounded by $\mathbf{E}\left[\norm{Q_{G}(\nabla\tilde{U}(\theta_k))}_1\right]\le G$,
\begin{align*}
    \mathbf{E}[\norm{\psi_k-\theta_k}_2^2] &=  \mathbf{E}\left[\norm{\left(\theta_{k-1} - \alpha  Q_{G}(\nabla\tilde{U}(\theta_{k-1}))\right) - \theta_{k} + \sqrt{2\alpha }\xi_{k}}_2^2\right]\\
    &= \mathbf{E}\left[\norm{\left(\theta_{k-1} - \alpha  Q_{G}(\nabla\tilde{U}(\theta_{k-1}))\right) - \theta_k }_2^2\right] +\mathbf{E}\left[\norm{ \sqrt{2\alpha }\xi_{k} }_2^2\right]\\
    &\le \max\left(2\mathbf{E}\left[\norm{\left(\theta_{k-1} - \alpha  Q_{G}(\nabla\tilde{U}(\theta_{k-1}))\right) - Q^s\left(\theta_{k-1} - \alpha  Q_{G}(\nabla\tilde{U}(\theta_{k-1}))\right) }_2^2\right], 4\alpha d\right).
\end{align*}
Using the bound equation (6) in \citet{li2019dimension} gives us, 
\begin{align*}
   &\hspace{-2em}\mathbf{E}\left[\norm{\left(\theta_{k-1} - \alpha  Q_{G}(\nabla\tilde{U}(\theta_{k-1}))\right) - Q^s\left(\theta_{k-1} - \alpha  Q_{G}(\nabla\tilde{U}(\theta_{k-1}))\right) }_2^2\right] \\
    &\le \Delta_W\alpha\mathbf{E}\left[\norm{Q_{G}(\nabla\tilde{U}(\theta_{k-1}))}_1\right]\\
    &\le \Delta_W\alpha G.
\end{align*}
It follows that
\begin{align*}
    \PExv{\norm{\psi_k-\theta_k}_2^2}
    &\le \max\left(2\Delta_W\alpha G, 4\alpha d\right).
\end{align*}
Let $A = \max\left(2\Delta_W\alpha G, 4\alpha d\right)$. The rest is that same as in the case $2\alpha > v_0$.

\end{proof}

\section{Comparison to SGD Bounds}\label{sec:bound_compare}

There have been many works on comparing optimization and sampling algorithms since they serve as two main computational strategies for machine learning~\citep{ma2019sampling,talwar2019computational}. For example, in~\citet{ma2019sampling}, the authors compare the total variation distance between the approximate distribution and the target distribution (sampling bound), with the objective gap  (optimization bound).
Following previous work, we compare our 2-Wasserstein distance bound with previous SGD bounds.
Previous low-precision SGD convergence bounds are shown in terms of the squared distance to the optimum $\norm{\bar{\theta}_K - \theta^*}_2^2$~\citep{yang2019swalp}. In order to compare our bounds with theirs, we consider a 2-Wasserstein distance between two point distributions. Let $\mu_K$ be the point distribution assigns zero probability everywhere except $\bar{\theta}_K $ and $\pi$ be the point distribution assigns zero probability everywhere except $\theta^* $. Then we get

\begin{align*}
    W_2(\mu_K, \pi) = \left(\inf_{J\in \mathcal{J}(x,y)}\int \norm{x-y}_2^2dJ(x,y)\right)^{1/2} \le \mathbf{E}\left[ \norm{ \bar{\theta}_K  - \theta^* }_2^2 \right]^{\frac{1}{2}}.
\end{align*}

From~\citet{yang2019swalp}, we know that $\mathbf{E}\left[ \norm{ \bar{\theta}_K  - \theta^* }_2^2 \right]^{\frac{1}{2}}$ is proportional to $\Delta_W$.
Therefore, our 2-Wasserstein distance is $O(\Delta_W^2)$ whereas SGD's 2-Wasserstein distance is $O(\Delta_W)$, which shows SGLD is more robust to the quantization error.

\section{Deterministic Rounding vs Stochastic Rounding }\label{sec:deterministic_rounding_compare}
Compared to deterministic rounding, stochastic rounding is unbiased and thus can preserve gradient information even when the gradient update is smaller than the quantization gap. In theory, deterministic rounding will make the convergence bound worse due to its bias. For example, in Theorem~\ref{thm:highacc}, using deterministic rounding as the weight quantizer makes the bias term becomes $\mathcal{O}(\Delta_w)$, which is worse than the current $\mathcal{O}(\Delta_w^2)$. In practice, stochastic rounding generally provides much better results than deterministic rounding especially on deep neural networks~\citep{gupta2015deep,wang2018training}. For example, on CIFAR10 with 8-bit block floating point, we found that using deterministic rounding to quantize the weight in SGDLP-L and SGLDLP-L gives test errors 7.44\% and 7.37\% respectively, which are much worse than using stochastic rounding (SGDLP-L:5.86\%, na\"ive SGLDLP-L: 5.85\%, VC SGLDLP-L: 5.51\%). 

\section{Algorithms with (Block) Floating Point Numbers}\label{sec:bfp}
The qunatization gaps in floating point and block floating point change depending on the number values. Therefore, we need to compute the qunatization gap in each step in order to apply our variance-vorrected quantization function $Q^{vc}$. It is easy to see that the qunatization gap can be computed as
\begin{equation}\label{eq:compute_delta}
      \Delta_W(\mu) \leftarrow
      \begin{cases}
      2^{E[\mu]-W+2} \text{ where } E[\mu]= \text{clip}(\left\lfloor\log_2(\max\Abs{\mu})\right\rfloor, l, u) &  \text{block floating point}\\
      2^{E[\mu]-W} \text{ where } E[\mu]= \text{clip}(\left\lfloor\log_2(\Abs{\mu})\right\rfloor, l, u) & \text{floating point.}
      \end{cases}
\end{equation}
Deterministic rounding and stochastic rounding are defined using the above $\Delta_W$. Then we obtain $Q^{vc}$ function with (block) floating point in Algorithm~\ref{alg:vc_fp}. This algorithm is the same as Algorithm~\ref{alg:vcsgld} except the lines in red where we recompute the quantization gap $\Delta$ after adding Gaussian noise to make sure it aligns with the quantization gap of $x$. VC SGLDLP-L with (block) floating point is outlined in Algorithm~\ref{alg:vcsgld_fp}.

\begin{algorithm}[H]
  \caption{VC SGLDLP-L with (Block) Floating Point.}
  \begin{algorithmic}
    \label{alg:vcsgld_fp}
    \STATE \textbf{given:} Stepsize $\alpha$, number of training iterations $K$, gradient quantizer $Q_G$, deterministic rounding with (block) floating point $Q^d$, stochastic rounding with (block) floating point $Q^s$, $F$ bits to represent the shared exponent (block floating point) or the exponent (floating point), $W$ bits to represent each number in the block (block floating point) or the mantissa (floating point).
    \STATE \textbf{let} $l \leftarrow -2^{F-1}, u \leftarrow 2^{F-1}-1$
      \FOR{$k = 1:K$}
      \STATE \textbf{compute} $\mu \leftarrow \theta_{k} - \alpha Q_G\left(\nabla\tilde{U}(\theta_{k-1})\right)$
      \STATE \textbf{compute} $\Delta_W(\mu)$ following Equation~\eqref{eq:compute_delta}
	\STATE \textbf{update} $\theta_{k+1} \leftarrow Q^{vc}\left(\mu, 2\alpha, \Delta_W(\mu)\right)$ 
    \ENDFOR
    \STATE \textbf{output}: samples $\{\theta_k\}$
   
  \end{algorithmic}
\end{algorithm}

\begin{algorithm}[H]
  \caption{Variance-Corrected Quantization Function $Q^{vc}$ with (Block) Floating Point.}
  \begin{algorithmic}\label{alg:vc_fp}
  \STATE \textbf{input}: ($\mu$, $v$, $\Delta$)
  \STATE $v_0\leftarrow\Delta^2/4$
    \IF{$v>v_0$}
    \STATE $x \leftarrow \mu + \sqrt{v-v_0}\xi$, where $\xi\sim\mathcal{N}(0,I_d)$
    \STATE $r \leftarrow x - Q^d(x)$ 
    {\color{red} \STATE recompute $\Delta \leftarrow \Delta_W(x)$ following Equation \eqref{eq:compute_delta}
    \STATE $v_0\leftarrow\Delta^2/4$
    }
    \FORALL{$i$}
    \STATE \textbf{sample} $c_i$ from Cat$(|r_i|,v_0)$ as in Equation~\eqref{eq:categorical}
    \ENDFOR
    \STATE $\theta\leftarrow Q^d(x)+\text{sign}(r)\odot c$
    \ELSE
    \STATE the same as in fixed point numbers
    \ENDIF
  \end{algorithmic}
\end{algorithm}

\section{Experimental Details and Additional Results}
\subsection{Sampling methods}
For both SGLD and low-precision SGLD, we collected samples $\{\theta_k\}_{K=1}^J$ from the posterior of the model’s weight, and obtained the prediction on test data $\{x^*,y^*\}$ by Bayesian model averaging
\[
p(y^*|x^*,\mathcal{D}) \approx \frac{1}{J}\sum_{j=1}^J p(y^*|x^*,\mathcal{D},\theta_j).
\]
\subsection{MNIST}
\begin{figure}[H]
	\vspace{-0mm}\centering
	\begin{tabular}{cccc}		
		\includegraphics[width=6cm]{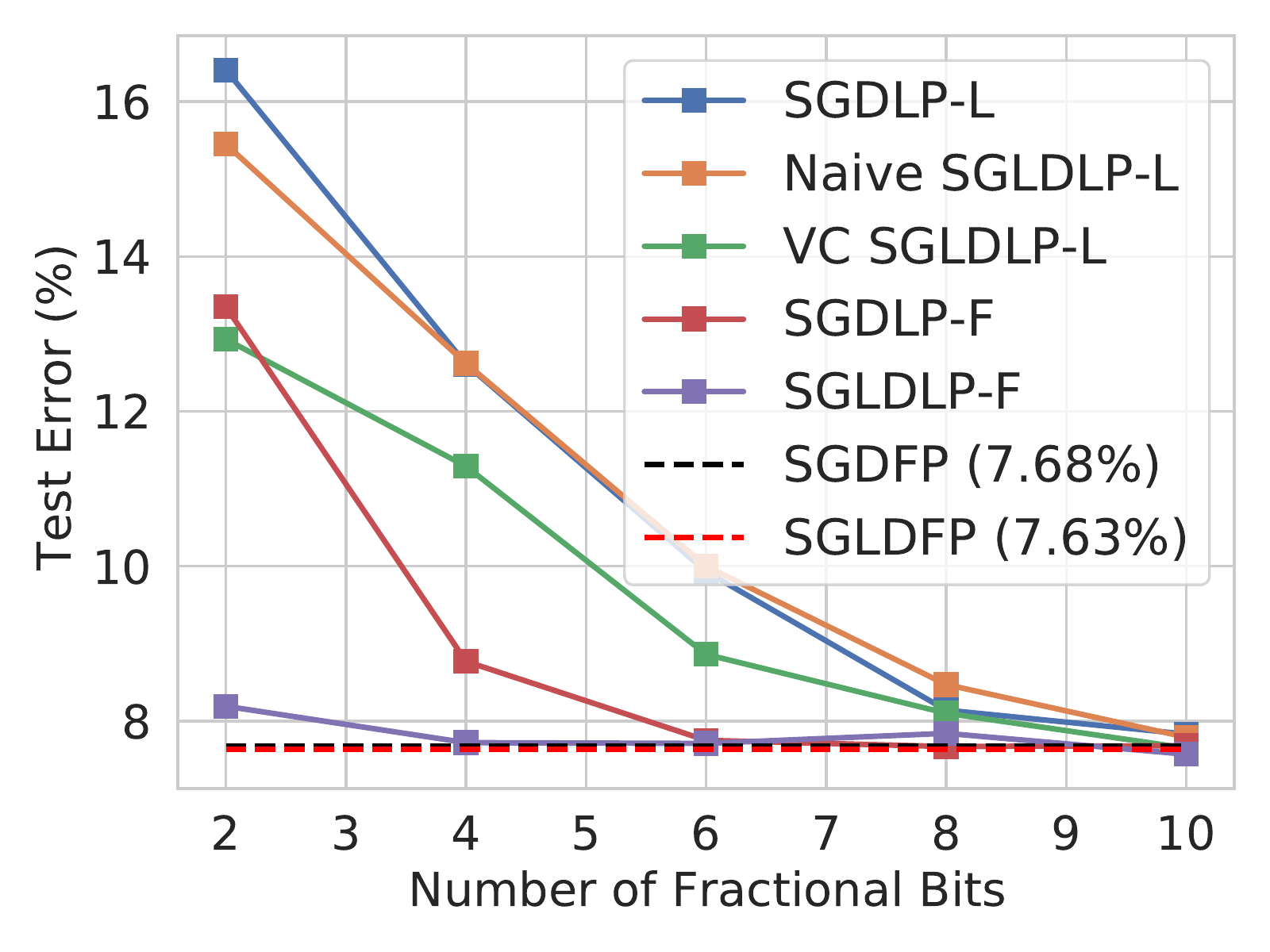}   &      
		\includegraphics[width=6cm]{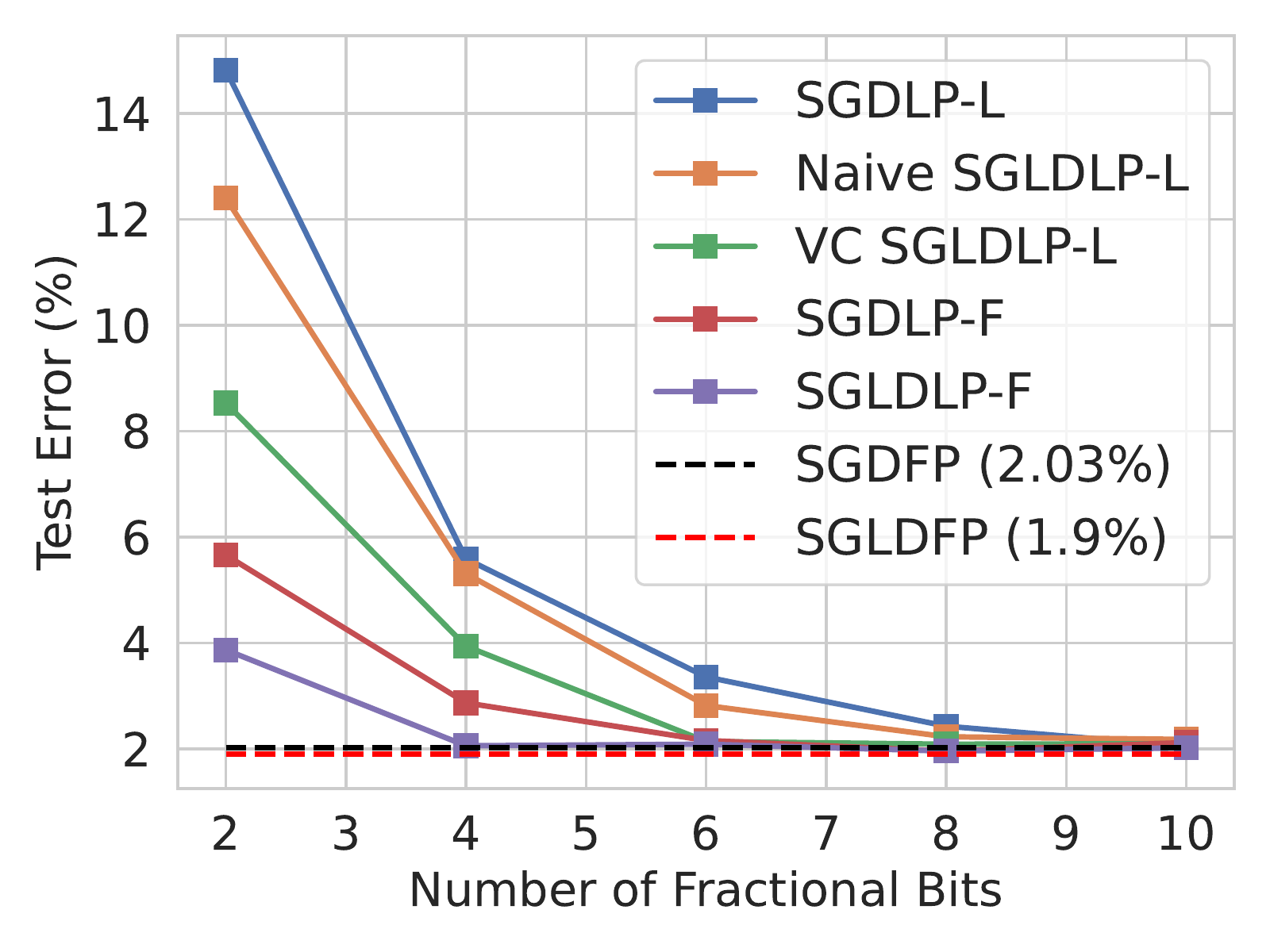}   \\
	\end{tabular}
	\caption{Test error on MNIST dataset in terms of different precision.}
	\label{fig:mnist_error}
\end{figure}

We train all methods on logistic regression and MLP for 20 epochs with learning rate 0.1 and batch size 64. We additionally report test error comparisons in Figure~\ref{fig:mnist_error}.

\subsection{CIFAR and IMDB}
For CIFAR datasets, we use batch size 128, learning rate $0.5$ and weight decay $5e-4$. We train the model for 245 epochs and used the same decay learning rate schedule as in~\citet{yang2019swalp}. We collect 50 samples for SGLD. For cyclical learning rate schedule, we use 7 cycles and collect 5 models per cycle (35 models in total). We use 10 bins for expected calibration error (ECE) following prior work~\citep{guo2017calibration}.

For IMDB dataset, we use batch size 80, learning rate $0.3$ and weight decay $5e-4$. We use a two-layer LSTM. The embedding dimension and the hidden dimension are 100 and 256 respectively. We train the model for 50 epochs and used the same decay learning rate schedule as on CIFAR datasets. We collect 20 samples for SGLD. For cyclical learning rate schedule, we use 1 cycles and collect 20 models.

\subsection{ImageNet}
We use batch size 256, learning rate $0.2$ and weight decay $1e-4$. We use the same decay learning rate schedule as in~\citet{he2016deep} and collect 20 models for SGLD.

\end{document}